\documentclass[12pt]{article}

\usepackage{amssymb}

\usepackage{times}
\usepackage{soul}
\usepackage{url}
\usepackage[hidelinks]{hyperref}
\usepackage[utf8]{inputenc}
\usepackage[small]{caption}
\usepackage{graphicx}
\usepackage{amsmath}
\usepackage{booktabs}
\usepackage{algorithm}
\usepackage{algorithmic}
\usepackage{latexsym}
\usepackage{amssymb}
\usepackage{epsfig}
\usepackage{subfigure}
\usepackage[table,xcdraw]{xcolor}
\usepackage{verbatim}

\newtheorem{mydef}{Definition}

\newcommand{\elpp}{{\cal{EL}^{++}}}
\newcommand{\elp}{{\cal {EL}^{+}}}
\newcommand{\el}{{\cal {EL}}}
\newcommand{\alc}{{\cal {ALC}}}
\newcommand{\bottom}{\bot}
\newcommand{\shoin}{{\cal{SHOIN}}}

\begin{document}




\title{Completing and Debugging Ontologies: \\ State of the Art and Challenges}


\author{Patrick Lambrix
\\
Link\"oping University, Sweden}


\date{}
\maketitle  

\begin{abstract}

As semantically-enabled applications require high-quality ontologies, developing and maintaining ontologies that are
as correct and complete as possible is an important although difficult task in ontology engineering. A key step is ontology debugging and completion. In general, there are two steps: detecting defects and repairing defects. In this paper we discuss the state of the art regarding the repairing step. We do this by formalizing the repairing step as an abduction problem and situating the state of the art with respect to this framework. We show that there are still many open research problems and show opportunities for further work and advancing the field.

\end{abstract}









\section{Introduction}
\label{sec:intro}

Ontologies (e.g., \cite{SS09}) aim to define the basic terms and relations of a domain of interest, as well as the rules for combining these terms and relations. They standardize terminology in a domain and are a basis for semantically enriching data, semantic search, integration of data from different data sources, and  reasoning over the data. Using ontologies can alleviate the variety (data sources are heterogeneous regarding the type and nature of data they store), variability (data can be inconsistent) and veracity (not all data can be trusted) problems. Furthermore, they are proposed as an enabler making data FAIR, i.e., findable, accessible, interoperable, and reusable, with the purpose of enabling machines to automatically find and use the data, and individuals to easily reuse the data \cite{wilkinson2016fair}. Ontologies are also a key technology for the semantic web.

In recent years many ontologies have been developed (see \cite{dN12} for a survey on ontology libraries). Further, ontologies have been connected to each other into ontology networks and there are some portals that store these (e.g., BioPortal (\url{http://bioportal.bioontology.org/}), Unified Medical Language System (\url{http://www.nlm.nih.gov/research/umls/about_umls.html})). However, developing ontologies and networks are not easy tasks and there may be issues related to the quality of the ontologies and networks \cite{SMB10}. Two such issues are incorrectness (does the ontology contain wrong information?) and completeness (is information lacking?).  Ontologies containing wrong information or lacking information, although often useful, also lead to problems when used in semantically-enabled applications. 
 Wrong conclusions may be derived or valid conclusions may be missed.  As an example, in \cite{LiL10} it was shown that semantically-enabled querying of PubMed (\url{http://www.ncbi.nlm.nih.gov/pubmed/}) using
MeSH (Medical Subject Headings, \url{http://www.nlm.nih.gov/mesh/}) with one piece of information missing (i.e., that scleritis is a scleral disease) would lead to missing 55\% of the result that is obtained with this piece of information. 
 Therefore, it is essential to complete and debug  ontologies and their networks.

Defects in ontologies can take different forms (e.g., \cite{KPSH05}). Syntactic defects are usually easy to find and to resolve. Defects regarding style include such things as unintended redundancy. More interesting and severe defects are the modeling defects and the semantic defects. Modeling defects relate to the domain that is being modeled and include such things as missing concepts and relations, or statements that are not correct according to the domain. 
For instance, in \cite{LLT09} it was shown 
that for the two ontologies used in the Anatomy track in the Ontology Alignment Evaluation Initiative (OAEI, yearly event for evaluation of ontology alignment systems), at least 121 and 83, respectively, is-a relations (defined below) that are correct in the domain are missing in these ontologies. 
Further,  BioPortal contains results for the OAEI Anatomy track with ca 20\% incorrect statements and ca 20\% of the connections between ontologies are missing.
Semantic defects relate to logical defects such as defining concepts that are logically equivalent to the empty set (called unsatisfiable concepts, formally defined below) or ontologies that contain contradictions. As an example, in \cite {KPSH05} it was shown that the TAMBIS ontology contained 144 unsatisfiable concepts.

In this paper we review approaches for raising the quality of ontologies by repairing them. In general, completing and debugging requires two steps. In the detection step possible defects are detected.
In the repairing step the detected wrong information is removed and missing information is added. In this paper we focus on the repairing step which can be formalized as an abduction problem and for which there are still many open research problems. 

The remainder of the paper is organized as follows. In Sect. \ref{sec-prel} we discuss preliminaries introducing notions related to ontologies (Sect. \ref{sec-ontologies}), description logics which are used as a basis for the formalization of the repairing problem (Sect. \ref{sec-dl}), and a short review of methods for the detection step (which is not the focus of the paper) and its relation to the repairing step (Sect. \ref{sec-workflow}). In Sect. \ref{sec-related-work} we relate this survey to earlier surveys.  Then, in Sect. \ref{sec-repair} we formalize ontology repair (completion and debugging) as an abductive reasoning problem and, as there may be different ways to repair ontologies, we introduce different preference relations between solutions that are relevant to this problem. In Sect. \ref{sec-state-of-the-art-ontologies} and \ref{sec-state-of-the-art-networks}  we discuss the state of the art based on our formalization for ontologies and ontology networks, respectively regarding debugging,  completion and the combination of these.  Further, we give some open problems related to theory and algorithms as well as regarding user involvement in Sect.  \ref{sec-opportunities}.
The paper concludes in Sect. \ref{sec-conclusion}.

\section{Preliminaries}
\label{sec-prel}

\subsection{Ontologies and ontology networks}
\label{sec-ontologies}

From a knowledge representation point of view, ontologies may contain four
components: (i) concepts that represent sets or classes of entities in a domain (e.g., Fracture in the example in the appendix, representing all fractures), (ii)  instances that represent the actual entities (e.g., an actual fracture), (iii) relations (e.g., hasAssociatedProcess in the example in the appendix), and (iv) axioms that represent facts that are always true in the topic area of the ontology (e.g., a fracture is a pathological phenomenon). Concepts and relations are often organized in hierarchies using the is-a (or subsumption) relation, denoted by $\sqsubseteq$. When $P$ $\sqsubseteq$ $Q$, then $P$ is a sub-concept of $Q$ and all entities belonging to $P$ also belong to $Q$. Axioms can represent such things as domain restrictions, cardinality restrictions, or disjointness restrictions. Many ontologies do not contain instances and represent knowledge on the concept (or 'schema') level. In this paper we do not deal with instances.

Ontologies can be represented in different ways, but one of the more popular ways nowadays is to use (variants of) the OWL language (e.g., \url{https://www.w3.org/TR/2012/REC-owl2-primer-20121211/}). OWL is based on description logics, which are presented in Sect. \ref{sec-dl}. In description logics concepts, roles, individuals and axioms are used, which relate to concepts, binary relations, instances and axioms in ontology terminology, respectively.

An ontology network is a collection of ontologies and pairwise alignments between these ontologies. An alignment is a set of mappings (also called correspondences) between entities from the different ontologies. The most common kinds of mappings are equivalence mappings as well as mappings using is-a and its inverse.

\subsection{Description logics}
\label{sec-dl}

In this paper we assume that ontologies are represented using a description logic TBox. We introduce notions in the field of description logics that are relevant for this paper.

Description logics \cite{BCMNP03} are knowledge representation languages. In description logics concept descriptions are constructed inductively from a set $N_C$ of atomic concepts and a set $N_R$ of atomic roles and (possibly) a set $N_I$ of individual names. 
Different description logics allow for different constructors for defining complex concepts and roles.
In current work on completing and debugging different logics are used such as $\el$ (which uses the top concept $\top$, and the concept constructors conjunction and existential restriction) and $\alc$ (top concept $\top$, bottom concept $\bottom$ and concept constructors conjunction, disjunction, negation, and existential  and universal restrictions).
We mention later also $\shoin$ which in addition to the $\alc$ constructors allows constructors for transitive roles, role hierarchies, nominals, inverse roles and number restrictions. In Table \ref{tab:alc} we show the syntax of the $\alc$ constructors and refer to, e.g., \cite{BHLS17}  for information on other description logics.

An interpretation $\cal I$ consists of a non-empty set $\Delta^{\cal I}$ and an interpretation function $\cdot^{\cal I}$ which assigns to each atomic concept $P_a \in N_C$ a subset $P_a^{\cal I} \subseteq \Delta^{\cal I}$, to each atomic role $r_a \in N_R$ a relation $r_a^{\cal I} \subseteq \Delta^{\cal I} \times \Delta^{\cal I}$, and to each individual name $i \in N_I$ an element $i^{\cal I} \in \Delta^{\cal I}$.
The interpretation function is straightforwardly extended to
complex concepts (see Table \ref{tab:alc}). 
A TBox is a finite set of \emph{general concept inclusions} (GCIs)  (in Table \ref{tab:alc}) and \emph{role inclusions} (RIs).
An interpretation ${\cal I}$ is a \emph{model} of a TBox $T$ if  for each GCI and RI in $T$, the semantic conditions are satisfied. 
We say that a TBox $T$ is \emph{inconsistent} if there is no model for $T$. Further, we say that a TBox is \emph{incoherent} if it contains an \emph{unsatisfiable} concept where a concept $P$ in a TBox $T$ is unsatisfiable if for all models  $\cal I$  of $T$: $P^{\cal I}$  = $\emptyset$. 

One of the main reasoning tasks for description logics is subsumption checking in which the problem is to decide for a TBox $T$ and concepts $P$ and $Q$ whether $T  \models P \sqsubseteq Q$, i.e., whether $P^{\cal I} \subseteq Q^{\cal I}$ for every model of Tbox $T$. We say then also that $Q$ \emph{subsumes} $P$ or that $P$ \emph{is-a} $Q$.

\begin{table}[]
\begin{center}
\begin{tabular}{|c|c|c|}
\hline
Name & Syntax & Semantics \\
\hline
\hline
top & $\top$ & $\Delta^{\cal I}$ \\
\hline
bottom & $\bottom$ & $\emptyset$ \\
\hline
negation & $\neg$ $P$  &  $\Delta^{\cal I}$ $\setminus$ $P^{\cal I}$ \\
\hline
conjunction & $P \sqcap Q$ & $P^{\cal I} \cap Q^{\cal I}$ \\
\hline
disjunction & $P \sqcup Q$ & $P^{\cal I} \cup Q^{\cal I}$ \\
\hline
existential restriction & $\exists r. P$ & $\{ x \in \Delta^{\cal I} ~ | \exists y \in \Delta^{\cal I} : $ $(x,y) \in r^{\cal I} \wedge y \in P^{\cal I}\}$\\
\hline
universal restriction & $\forall r. P$ & $\{ x \in \Delta^{\cal I} ~ | \forall y \in \Delta^{\cal I} : $ $(x,y) \in r^{\cal I} \rightarrow y \in P^{\cal I}\}$\\
\hline
\hline
GCI & $P \sqsubseteq Q$ & $P^{\cal I} \subseteq Q^{\cal I}$ \\
\hline
\end{tabular}
\end{center}
\caption{$\alc$ syntax and semantics.}
\label{tab:alc}
\end{table}

\subsection{Completing and debugging workflow}
\label{sec-workflow}

In this paper we review approaches for raising the quality of ontologies by repairing them. In general, completing and debugging requires two steps. In the detection step defects are found using different approaches.
In the repairing step, on which this paper focuses, the detected wrong information is removed and missing information is added. Although we focus on the repairing step and most work on repairing assumes that the detection step is done, we briefly discuss detection and the connection to repairing.

\subsubsection{Detection}

There are many kinds of approaches to detect defects in ontologies and many of these are complementary.

A detection method that works for all kinds of defects is inspection of the ontologies. This requires ontology development environments that provide search, reasoning and explanation facilities to aid the domain expert in the inspection.

Most detection methods for semantic defects are logic-based and focus on wrong information in the ontologies. A common
strategy is to detect unsatisfiable concepts or inconsistencies in
ontologies or ontology networks using standard reasoning techniques. One problem that is reported in \cite{SC03} is that, as an unsatisfiable concept that is used in the definition of other concepts may make these unsatisfiable as well,  description logic reasoners could give large lists of unsatisfiable concepts. One way to alleviate this problem is as in \cite{KPSH05} where these 'root' concepts are identified (although in \cite{KPSH05} they do this during the repairing step).

There are many approaches to find missing information.
There is much work on finding candidate relationships between terms in the ontology learning area \cite{BCM05}. In this setting, new ontology elements are derived from text using knowledge acquisition techniques. Another paradigm is based on machine learning and statistical methods, such as k-nearest neighbors approach \cite{MPS03}, association rules \cite{MS00}, bottom-up hierarchical clustering techniques \cite{ZPVP07}, supervised classification \cite{SVK10} and formal concept analysis \cite{CHS05}.

A much used approach is to use patterns.
The pioneering research conducted in this line is in \cite{Hearst92}, where the focus was on finding missing is-a relations. The work defines a set of lexico-syntactical patterns indicating is-a relationships between words in the text. However, depending on the chosen corpora, these patterns may occur rarely. Thus, although the approach has a reasonable precision, its recall is very low. Many variants have been proposed. 
Lexico-syntactic patterns as well as logic patterns have been used to find wrong as well as missing information in ontologies (e.g., \cite{CRVP09,Keet12,RZ13,PGS14}) and ontology networks (e.g., \cite{WX08,BH12}). The OOPS! system implements a variety of these \cite{PGS14}.

There are also approaches that use knowledge that is intrinsic in an ontology network to detect defects. For instance, in \cite{IL13,LI13} a partial alignment between ontologies is used to detect missing is-a relations. These are found by looking at pairs of equivalence mappings. If there is an is-a relation between the terms in the mappings belonging to one ontology, but there is no is-a relation between the corresponding terms in the other ontology, then it is concluded that there is a candidate missing is-a relation in the second ontology. A similar approach is used in \cite{BH08}.


The detection of missing mappings is a research area on its own, i.e. ontology alignment \cite{ES07}, and we discuss this further in Sect. \ref{sec-state-of-the-art-networks}.

We note that for missing information these detection approaches usually do not detect {\it all} defects and they do not  guarantee that the found defects are really missing. The found defects are really {\it candidate} defects which need to be validated by a domain expert.

\subsubsection{Workflow}

In much of the current work we find systems or methods that detect defects or repair defects, but usually not both. 

A workflow for a system for completing and debugging ontologies and ontology networks contains two main steps: detection and repair. It is well-known that for high-quality results a domain expert needs to be involved in both steps. As the defects found by detection systems usually are candidate defects, a domain expert needs to validate the candidate defects as wrong input to the repairing step would lead to wrong repairs of the ontologies. Furthermore, in the repairing step domain experts are needed to validate repairs for modeling problems. Also for the semantic defects a domain expert is needed, as systems that are purely logic based may prefer logically correct solutions that are not correct according to the domain over solutions that are (e.g., \cite{PFSC13}).
The two steps do not need to be completely separated. 
For instance, when repairing the ontology new information is added or wrong information is deleted and this may be used to detect further defects. 

As an example, the RepOSE system \cite{IL13,LI13} is a system for debugging and completing is-a structure and mappings in ontology networks, where only the named concepts and the is-a relations in the ontologies are considered. RepOSE has a detection step that uses knowledge intrinsic in the network to detect candidate missing is-a relations and mappings. These candidate defects are then validated by a domain expert. In the case a candidate missing is-a relation or mapping really is missing, then we need to complete, otherwise a wrong is-relation or mapping is derivable from the network and thus we need to debug. The validated and classified defects (missing or wrong) are then used as input for the repairing step. Different algorithms are used for repairing different kinds of defects, but the sub-steps for each kind are generation of repairing actions (what to add or delete), the ranking of repairs (a proposed order in which to deal with the defects), the recommendation of repairing actions (using external knowledge) and finally, the execution of  the repairing actions chosen by the domain expert (with computation of the consequences of the action). The consequences can include such things as other defects are also repaired, possible repairing actions for other defects change or new candidate defects are found.
Furthermore, at any time during the process, the user can switch between different ontologies, start earlier phases, or switch between the repairing of different kinds of defects. The process ends when there are no more defects to deal with.

\section{Related work}
\label{sec-related-work}

There are early surveys from 2007 \cite{BQL07,HQ07} where debugging approaches are reviewed. In this paper we introduce a framework with preference relations that in addition to debugging also includes completion, and that allows us to compare the different approaches in a uniform way. The early surveys discuss approaches where the ontologies can include instances, which we do not. Further, \cite{BQL07} introduces some criteria for debugging approaches. Regarding the criterion \emph{application} the authors distinguish between different tasks. In this paper we focus on the repair task. The \emph{granularity} of the repairs is on the axiom level in this paper. Regarding \emph{Tbox and Abox support}, we focus on TBox support. Some algorithms focus on \emph{consistency} and some on \emph{incoherence}. For some of the methods, an \emph{implementation} is available. Regarding \emph{support of ontology networks}, we have a dedicated section to this topic (Sect. \ref{sec-state-of-the-art-networks}). \emph{User involvement} is discussed in Sect. \ref{sec-influence-oracle} and Sect. \ref{sec-user-involvement}. We mention the criteria \emph{preservation of structure}, \emph{complexity}  and \emph{exploitation of context} in relevant places in this paper.

\section{Ontology Repair}
\label{sec-repair}

In this section we focus on repairing ontologies represented in description logics, where we have already detected wrong and missing information. We only discuss ontologies at concept level, and thus do not deal with instances (or individuals in description logic terminology).
We define this problem as an abductive reasoning problem, and as a repairing problem can have many solutions, we discuss preference relations between these.
Further, we use an abstract example to exemplify the notions. However, in the appendix we give an example inspired by the Galen ontology.

\subsection{Formalization}
\label{sec-repair-formalization}

\subsubsection{Repair}

\begin{mydef} (Repair)
Let $T$ be a TBox and $C$ be the set of all atomic concepts in $T$. 
Let $M$ and $W$ be finite sets of TBox axioms.
Let $Or$ be an oracle that given a TBox axiom returns true or false.
A repair for Complete-Debug-Problem CDP$(T, C, Or, M, W)$ is any pair of finite sets of TBox axioms $(A,D)$ such that \\
(i) $\forall$ $\psi_a$ $\in$ $A$: $Or$($\psi_a$) = true;\\
(ii) $\forall$ $\psi_d$ $\in$ $D$: $Or$($\psi_d$) = false; \\
(iii) $(T \cup A) \setminus D$ is consistent;\\
(iv) $\forall$ $\psi_m$ $\in$ $M$: $(T \cup A) \setminus D$ $\models$ $\psi_m$;\\
(v) $\forall$ $\psi_w$ $\in$ $W$: $(T \cup A) \setminus D$ $\not \models$ $\psi_w$. 
\label{Def-repair-ontology}
\end{mydef}


Def. \ref{Def-repair-ontology} formalizes the repair of an ontology for which missing and wrong information is given. An ontology is represented by a TBox $T$ with its set of atomic concepts $C$. The identified  missing and wrong information is represented by a set $M$ of missing axioms, and a set $W$ of wrong axioms. To repair the TBox, a set $A$ of axioms that are correct according to the oracle should be added to the TBox and a set $D$ of axioms that are not correct according to the oracle should be removed from the TBox such that the new TBox is consistent, the missing axioms are derivable from the new TBox and the wrong axioms are not derivable from the new TBox. 
As an example, consider the CDP in Fig. \ref{CDP-example} and visualized in Fig. \ref{CDP-example-graph}. Then $R_1$, $R_2$, $R_3$, $R_4$ and $R_5$ (visualized in Fig. \ref{fig:CDP-example-repairs}) are all repairs of the CDP.

\begin{figure*}[tb]
\begin{center}
\begin{tabular}{| l |} \hline
T: \{ax1: $P_1 \sqsubseteq P_2$, 
ax2: $P_1 \sqsubseteq P_3$,  
ax3: $P_1 \sqsubseteq \neg P_4$,
ax4: $P_2 \sqsubseteq P_4$,  
ax5: $P_2 \sqsubseteq P_5$,  \\
\hspace{0.6cm} ax6: $P_3 \sqsubseteq P_5$,  
ax7: $P_3 \sqsubseteq P_6$,  
ax8: $P_4 \sqsubseteq P_7$,  
ax9: $P_5  \sqsubseteq \forall s.P_8$,  \\
\hspace{0.6cm} ax10: $P_6 \sqsubseteq \exists s.\neg P_8$\} \\
C: \{$P_1$, $P_2$, $P_3$, $P_4$, $P_5$, $P_6$, $P_7$, $P_8$\}\\
Or(X) = true for X = 
$P_1 \sqsubseteq P_3$ (ax2), 
$P_1 \sqsubseteq \neg P_4$ (ax3), 
$P_1 \sqsubseteq P_6$, 
$P_2 \sqsubseteq P_3$,
\\
\hspace{0.6cm}  
$P_2 \sqsubseteq P_4$ (ax4),
$P_2 \sqsubseteq P_5$ (ax5), 
$P_2 \sqsubseteq P_6$, 
$P_2 \sqsubseteq P_7$, 
$P_2 \sqsubseteq \forall s.P_8$,
\\
\hspace{0.6cm}  
$P_3 \sqsubseteq P_6$ (ax7), 
$P_4 \sqsubseteq P_3$,
$P_4 \sqsubseteq P_5$, 
$P_4 \sqsubseteq P_6$,
$P_4 \sqsubseteq P_7$ (ax8), 
\\
\hspace{0.6cm}  
$P_4 \sqsubseteq \forall s.P_8$,
$P_5  \sqsubseteq \forall s.P_8$ (ax9), 
$P_7 \sqsubseteq P_3$, 
$P_7 \sqsubseteq P_6$,
\\
\hspace{0.6cm}  
and axioms derivable from this list (e.g.,  if P $\sqsubseteq$ Q, then also  P $\sqcap$ O $\sqsubseteq$ Q.) \\
Or(X) = false if X is not in the list above (for true) or cannot be derived from  \\ the axioms in the list above. \\
M = \{$P_4 \sqsubseteq P_5$\} \\
W = \{$P_1 \sqsubseteq \bottom$, $P_3 \sqsubseteq \bottom$\}\\

\hline

$R_1$=(\{$P_4 \sqsubseteq P_5$, $P_7 \sqsubseteq P_3$\}, \{$P_1 \sqsubseteq P_2$ (ax1), $P_3 \sqsubseteq P_5$ (ax6), $P_6 \sqsubseteq \exists s.\neg P_8$ (ax10)\})\\
$R_2$=(\{$P_4 \sqsubseteq P_5$, $P_7 \sqsubseteq P_3$\}, \{$P_1 \sqsubseteq P_2$ (ax1), $P_6 \sqsubseteq \exists s.\neg P_8$ (ax10)\})\\
$R_3$=(\{$P_7 \sqsubseteq P_3$\}, \{$P_1 \sqsubseteq P_2$ (ax1), $P_6 \sqsubseteq \exists s.\neg P_8$ (ax10)\})\\
$R_4$=(\{$P_4 \sqsubseteq P_5$\}, \{$P_1 \sqsubseteq P_2$ (ax1), $P_3 \sqsubseteq P_5$ (ax6)\})\\
$R_5$=(\{$P_4 \sqsubseteq P_5$, $P_7 \sqsubseteq P_3$\}, \{$P_1 \sqsubseteq P_2$ (ax1), $P_3 \sqsubseteq P_5$ (ax6)\})\\

\hline
\end{tabular}
\end{center}
\caption{Example complete-debug problem.}
\label{CDP-example}
\end{figure*}

 \begin{figure}[tb] 
\centering
\includegraphics[width=1\textwidth]{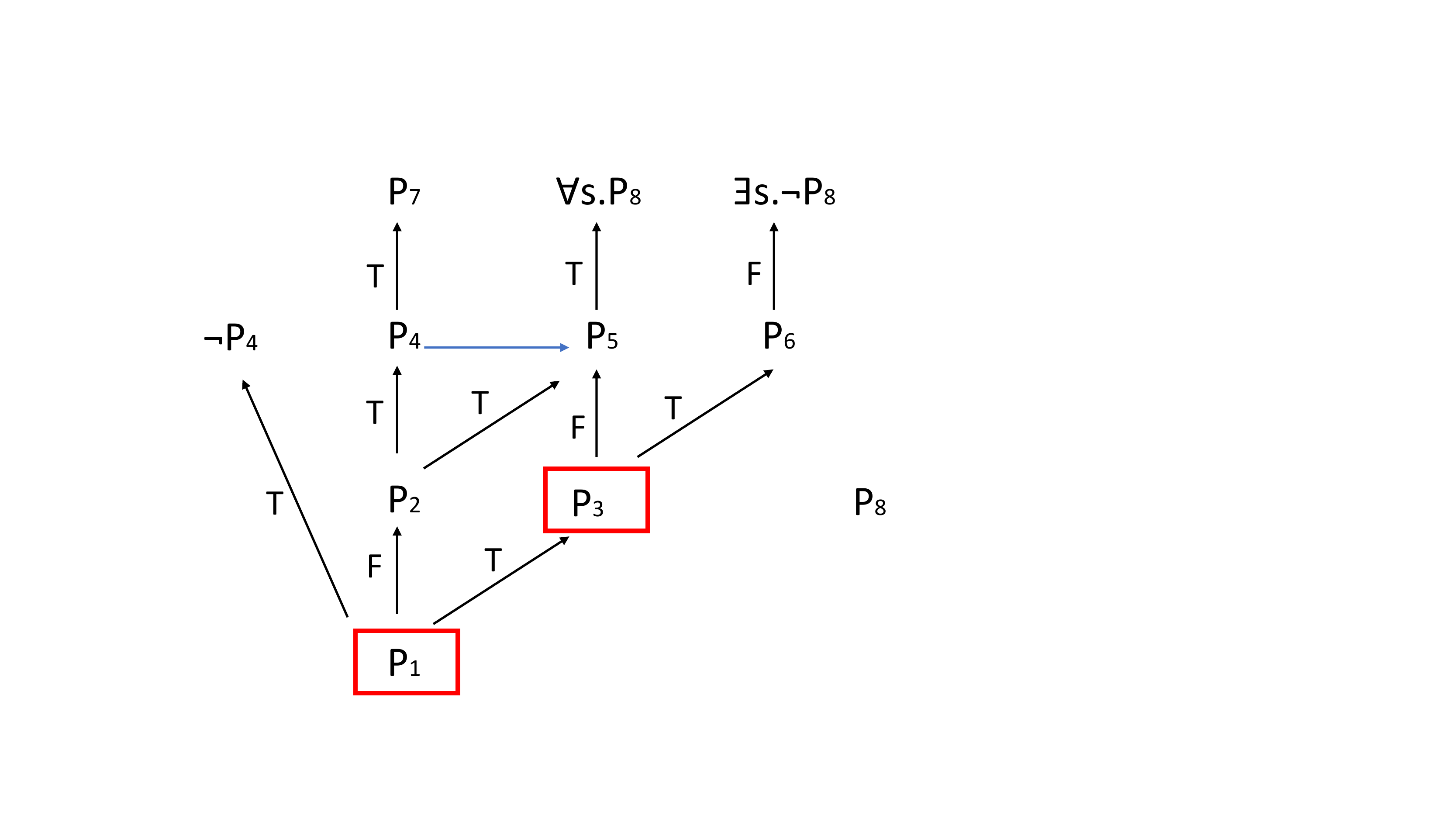}
\caption{Visualization of the example complete-debug problem in Fig. \ref{CDP-example}. The axioms in the Tbox are represented with black arrows. The missing axiom is represented in blue. The unsatisfiable concepts are marked with a red box. The oracle's knowledge about the axioms in the ontology is marked with T (true) or F (false) at the arrows.}
\label{CDP-example-graph}
\end{figure}

	\begin{figure}[t]
  \centering
  \subfigure[]{
    \includegraphics[trim=0mm 6mm 0mm 8mm, width=0.48\textwidth]{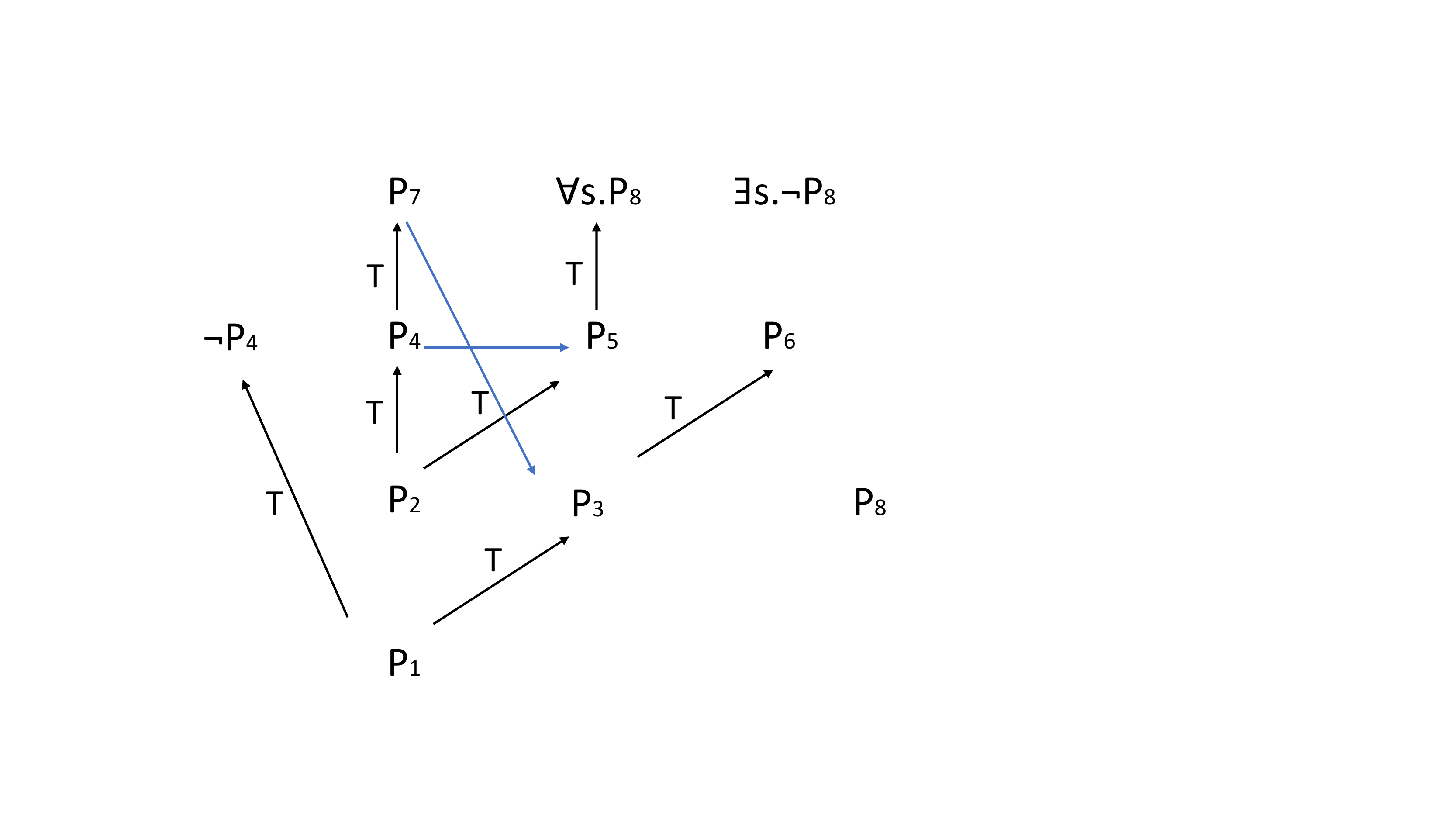}}
  \subfigure[]{
    \includegraphics[trim=0mm 6mm 0mm 8mm, width=0.48\textwidth]{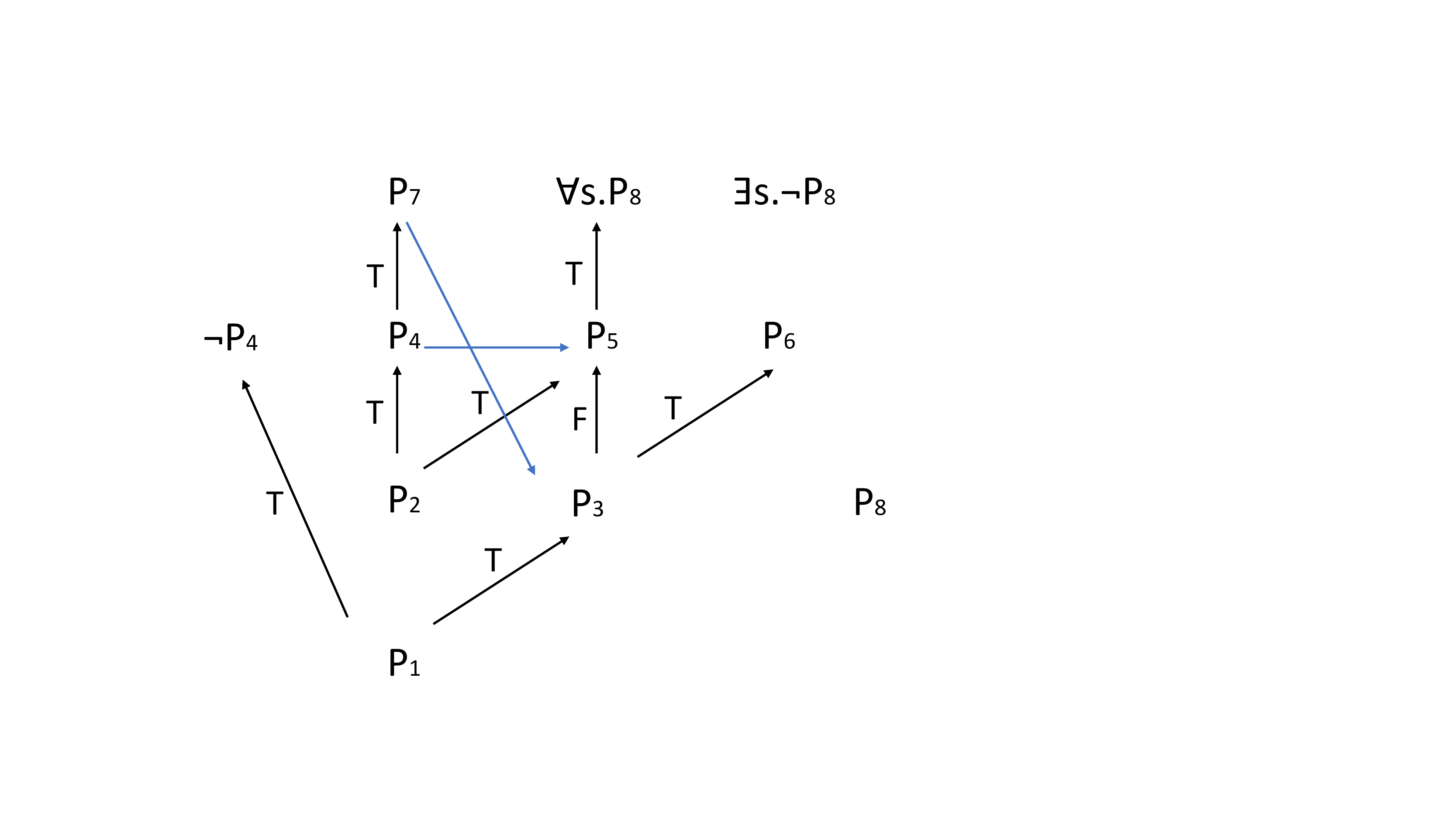}}
     \subfigure[]{
    \includegraphics[ trim=0mm 6mm 0mm 8mm, width=0.48\textwidth]{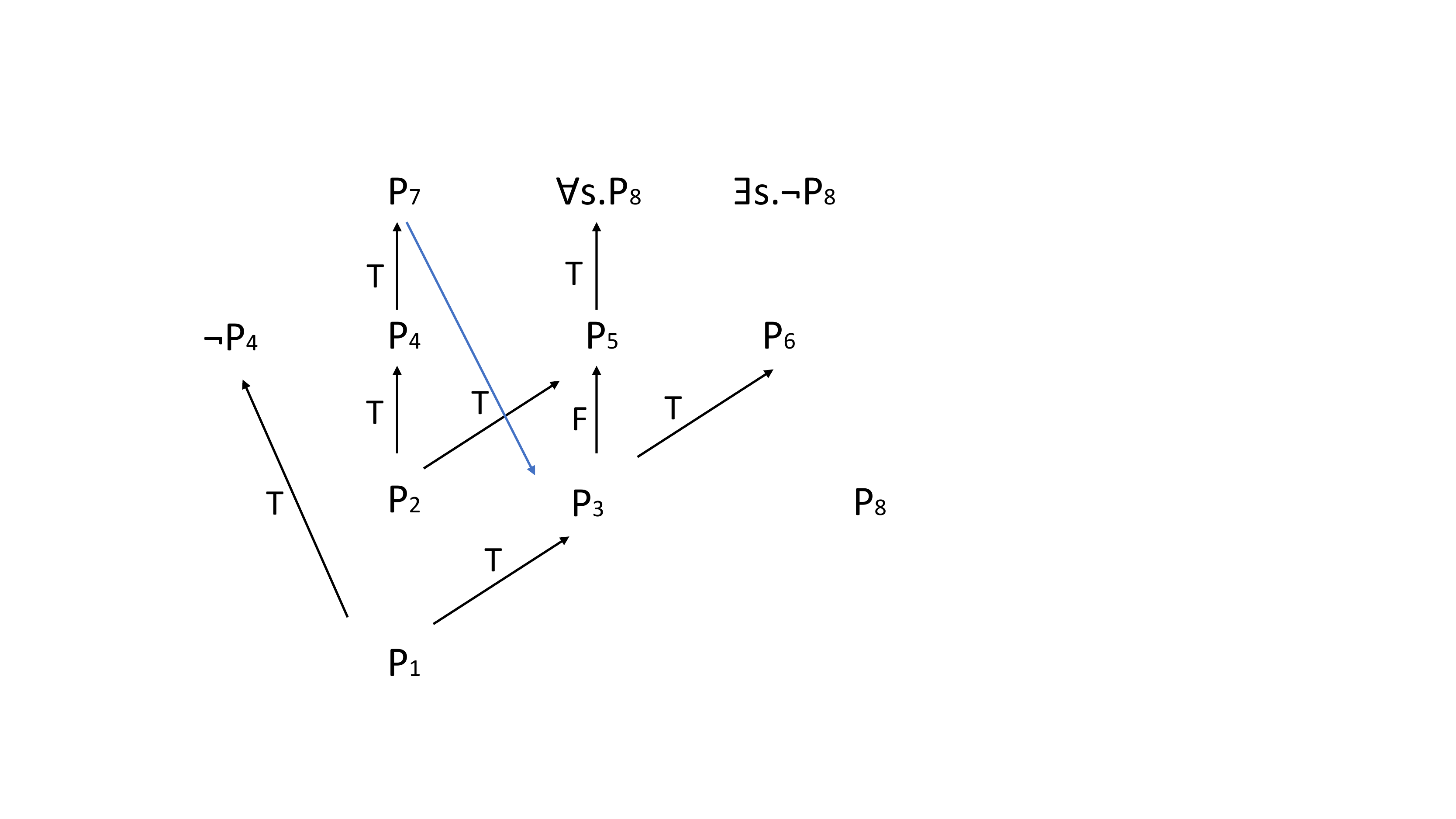}}
    \subfigure[]{
    \includegraphics[ trim=0mm 6mm 0mm 8mm, width=0.48\textwidth]{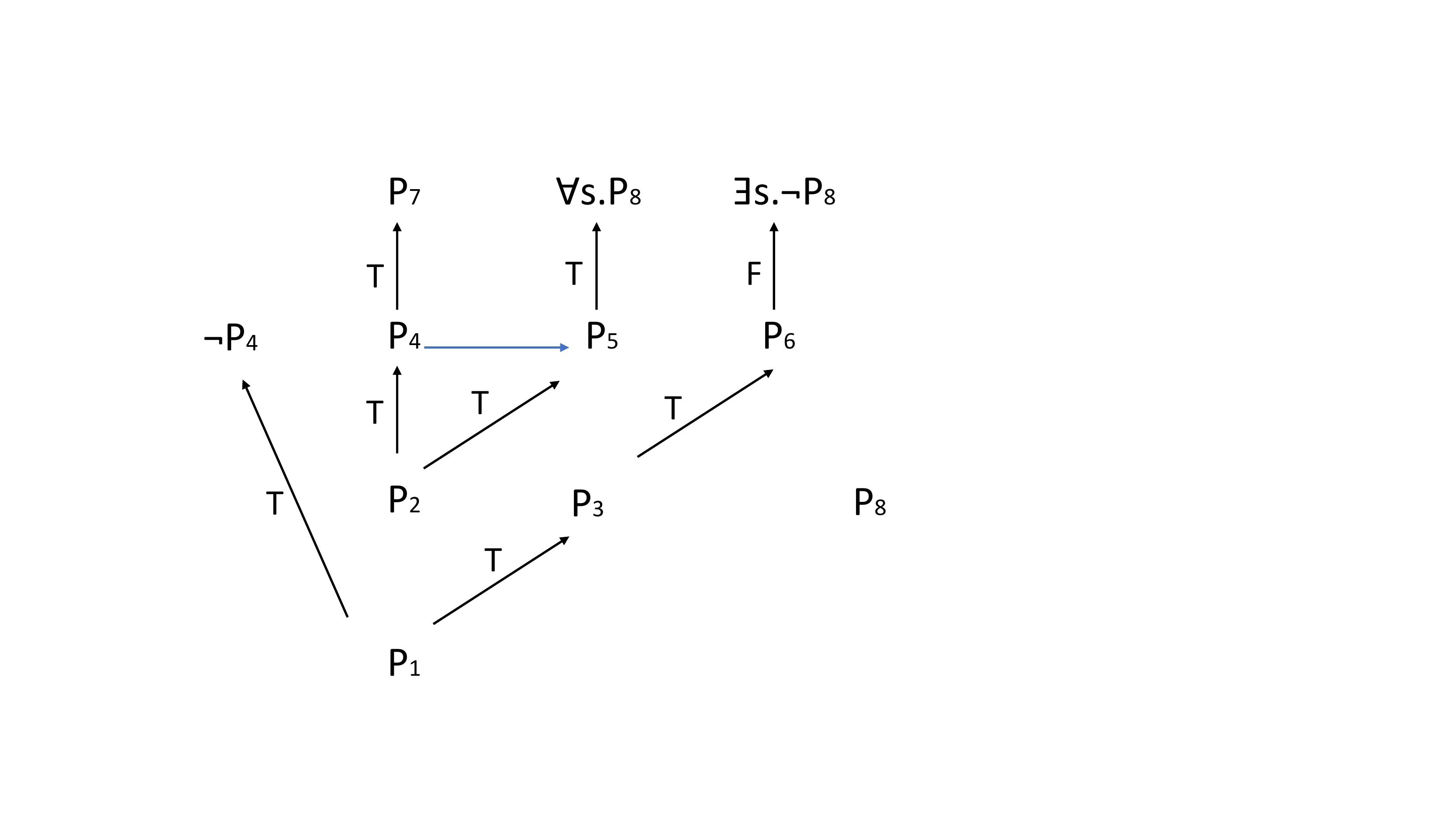}}
    \subfigure[]{
    \includegraphics[ trim=0mm 6mm 0mm 8mm, width=0.48\textwidth]{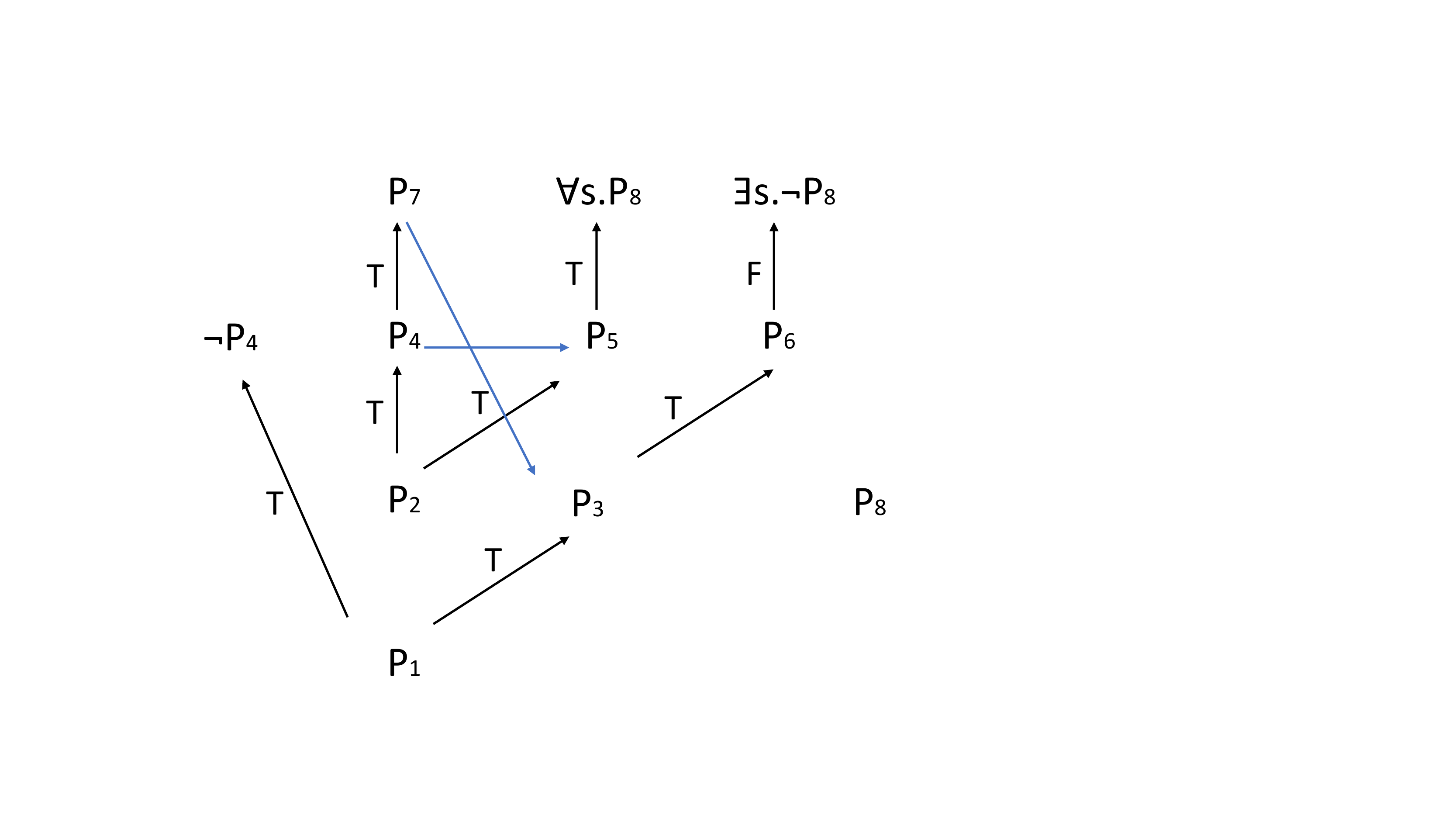}}
  \vspace{-10pt}
    \caption{Repairs (a) R1, (b) R2, (c) R3, (d) R4, (e) R5 for the example in Fig. \ref{CDP-example}.}  \label{fig:CDP-example-repairs}
  \vspace{-6pt}
\end{figure}

In general, the set of all axioms that are correct according to the domain and the set of all axioms that are not correct according to the domain are not known beforehand. Indeed, if these sets were given then we would only have to add the axioms of the first set to the TBox and remove the axioms in the second set from the TBox. The common case, however, is that we do not have these sets, but instead, we can rely on a domain expert who can decide whether an axiom is correct or not according to the domain. Therefore, in the formalization we introduce an oracle $Or$ that represents the domain expert and that when given an axiom, returns true or false.

\subsubsection{Influence of the quality of the oracle.}
\label{sec-influence-oracle}

For $Or$ we identified the following interesting cases. 
The first case is the {\it all-knowing oracle}. In this case the oracle's answer is always correct. This is the ideal case, but may not always be achievable. Most current work considers this kind of oracle.
In the second case, the {\it limited all-knowing oracle}, if $Or$ answers, then the answer is correct, but it may not know the answer to   all questions. This case represents a domain expert who knows a part of the domain well. 
An approximation of this case is when there are several domain experts who may have different opinions and we use a skeptical approach. Only if all domain experts give the same answer regarding the correctness of an axiom, do we consider the answer. 
In the third case $Or$ can make mistakes regarding the validation of axioms. Axioms that are not correct according to the domain may be validated as correct and vice versa. This is the most common case. 
Although most current work assumes an all-knowing oracle, recent work used,  in addition to an all-knowing oracle, also oracles with specific error rates in the evaluation of ontology alignment systems \cite{DILFJP16}. A lesson learned was that oracles with error rates up to 30\% were still beneficial for the systems.
The fourth case represents situations where no domain expert is available and there is no validation of axioms, such as in fully automated systems. 

As noted, most current work considers an all-knowing oracle. With an all-knowing oracle we can check that  $\forall$ $\psi_m$ $\in$ $M$: Or($\psi_m$) = true, and $\forall$ $\psi_w$ $\in$ $W$: Or($\psi_w$) = false and if this is not the case, we can remove the falsely identified defects. Therefore, we can, without loss of generality, assume that the axioms in $M$ really are missing, and the axioms in $W$ really are false. Furthermore, regarding repairs, when using an all-knowing oracle, we know that all added axioms in $A$ are correct according to the domain and all removed axioms in $D$ are false according to the domain. Furthermore, for an all-knowing oracle, we know that $A~\cap~ D = \emptyset$ (and then $(T \cup A) \setminus D$ = $(T \setminus D) \cup A$). When using other oracles, we cannot be sure that the given missing and wrong axioms really are missing and wrong, respectively. Therefore, oracles that make mistakes or do not know the correctness of all axioms may start with wrong input. Also, wrong axioms may be added and correct axioms may be removed during the repairing. These issues may have a negative effect on the quality of the repaired ontology.

In practice, when using domain experts, it is not possible to know which kind of domain expert is used. When only one domain expert is available it is reasonable for the systems to assume that an all-knowing expert is used, although we should be aware that mistakes can occur. When more domain experts are available, a skeptical approach or a voting approach may be used for raising the quality of the ontology.

\subsection{Preference relations}

As there may exist many possible repairs for a given CDP, and not all are equally interesting, it is necessary to define preference relations between repairs. 

\subsubsection{Basic preferences}

From the completion perspective of a complete-debug-problem it is important to find repairs that add as much correct information as possible to the ontology, while from the correctness perspective wrong information should be removed as much as possible. Def. \ref{def-more-complete} and \ref{def-more-correct} formalize these intuitions respectively. 

Def. \ref{def-more-complete} states that a repair $R$ is more complete than another repair $R'$ if all correct statements that can be derived from the ontology repaired by $R'$ also can be derived from the ontology repaired by $R$ and there is a correct statement that can be derived from the  ontology repaired by $R$, but not from the ontology repaired by $R'$. Therefore, if $S$ is more complete than another repair $R'$ then the ontology repaired by $R$ contains more correct statements than the ontology repaired by $R'$. 

Further, when the same correct statements can be derived from the ontology repaired by $R$ and the ontology repaired by $R'$, then $R$ and $R'$ are equally complete.

\begin{mydef} (more complete)
\label{def-more-complete}
Let $R=(A,D)$ and $R'=(A',D')$ be two repairs for CDP$(T, C, Or, M, W)$.
$R$ is \emph{more complete} than $R'$  (or $R$ is preferred to $R'$ with respect to 'more complete') iff \\
$(\forall \psi: ((T \cup A') \setminus D' \models \psi \wedge ~Or(\psi) = true )  \rightarrow  (T \cup A) \setminus D \models \psi)) \\
\wedge (\exists \psi:  Or(\psi) = true ~\wedge (T \cup A) \setminus D \models \psi  \wedge (T \cup A') \setminus D' \not\models \psi)$.\\
$R$ and $R'$ are \emph{equally complete} iff \\
$\forall \psi: Or(\psi) = true  \rightarrow 
((T \cup A') \setminus D' \models \psi  \leftrightarrow  (T \cup A) \setminus D \models \psi))$
\end{mydef}

Def. \ref{def-more-correct} states that a repair $R$ is less incorrect than another repair $R'$ if all wrong statements that can be derived from the ontology repaired by $R$ also can be derived from the ontology repaired by $R'$ and there is a wrong statement that can be derived from the ontology repaired by $R'$, but not from the ontology repaired by $R$. Therefore, if $R$ is less incorrect than another repair $R'$ then the ontology repaired by $R$ contains less wrong statements than the ontology repaired by $R'$. 

Further, when the same wrong statements can be derived from the ontology repaired by $R$ and the ontology repaired by $R'$, then $R$ and $R'$ are equally incorrect.

\begin{mydef} (less incorrect)
\label{def-more-correct}
Let $R=(A,D)$ and $R'=(A',D')$ be two repairs for CDP$(T, C, Or, M, W)$.
$R$ is \emph{less incorrect} than $R'$  (or $R$ is preferred to $R'$ with respect to 'less incorrect') iff \\
$(\forall \psi: ((T \cup A) \setminus D \models \psi \wedge Or(\psi) = false )   \rightarrow  (T \cup A') \setminus D' \models \psi)) \\ 
\wedge (\exists \psi:  Or(\psi) = false ~\wedge (T \cup A) \setminus D \not\models \psi  \wedge (T \cup A') \setminus D' \models \psi)$.\\
$R$ and $R'$ are equally incorrect iff \\
$\forall \psi: Or(\psi) = false  \rightarrow 
((T \cup A') \setminus D' \models \psi  \leftrightarrow  (T \cup A) \setminus D \models \psi))$
\end{mydef}

Def. \ref{def-subset} defines a classical preference relation for abduction problems related to removing redundancy using the subset relation. It compares the add and delete sets of two repairs.

\begin{mydef} (subset) 
\label{def-subset}
Let $R=(A,D)$ and $R'=(A',D')$ be two repairs for CDP$(T, C, Or, M, W)$. \\
$R$ $\subset_A$ $R'$ iff $A$ $\subset$ $A'$.\\
$R$ $\subset_D$ $R'$ iff $D$ $\subset$ $D'$. \\
$R$ $\subset$ $R'$ iff $R$ $\subset_A$ $R'$ $\wedge$ $R$ $\subset_D$ $R'$.\\
(If  $R$ $\subset$ $R'$, we also say that $R$ is preferred to $R'$ with respect to $\subset$.)
\end{mydef}

As examples, for the CDP in Fig. \ref{CDP-example} we have that $R_3$  $\subset$ $R_2$  $\subset$ $R_1$  and $R_4$  $\subset$ $R_5$ $\subset$ $R_1$.
Further, $R_1$ is less incorrect than $R_2$, $R_3$, $R_4$ and  $R_5$. $R_2$ and $R_3$ are equally incorrect, and $R_4$ and $R_5$ are equally incorrect.
We also have that $R_1$, $R_2$, $R_3$ and $R_5$ are equally complete and they are more complete than $R_4$.

\subsubsection{Preferred repairs with respect to a basic preference} 

Based on these preference relations we can define repairs that are preferred with respect to one particular preference relation (Def. \ref{def-max-complete} - \ref{def-min}). 

\begin{mydef} (maximally complete)
\label{def-max-complete}
A repair $R=(A,D)$ for CDP$(T, C, Or, M, W)$ is said to be maximally complete (or preferred with respect to 'more complete') iff there is no repair $R'$ which is more complete than $R$. 
\end{mydef}

\begin{mydef} (minimally incorrect)
\label{def-max-correct}
A repair $R=(A,D)$ for CDP$(T, C, Or, M, W)$ is said to be minimally incorrect (or preferred with respect to 'less incorrect') iff there is no repair $R'$ which is less incorrect than $R$.
\end{mydef}

\begin{mydef} (subset minimal)
\label{def-min}
A repair $R$ for CDP$(T, C, Or, M, W)$  is said to be subset minimal (or preferred with respect to $\subset$) iff there is no repair $R'$ such that $R'$ $\subset_A$ $R$ and $R'$ $\subset_D$ $R$. 
\end{mydef}

As examples, for the CDP in Fig. \ref{CDP-example} we have that $R_3$ and $R_4$ are subset minimal,
$R_1$ is minimally incorrect, and
$R_1$, $R_2$, $R_3$ and $R_5$ are maximally complete.

\subsubsection{Combining preferences} 

The criteria regarding completeness and correctness are desirable as completeness leads to more correct information and correctness leads to less incorrect information. In most cases also the reduction of redundancy is desirable (but see below for cases where this is not the case).
Therefore, we define different ways to combine these criteria. First, we need to define when a repair dominates another repair with respect to preference relations (Def. \ref{def-dominate}). A repair $R$ dominates  another repair $R'$ if $R$ is at least equally preferred to $R'$ for each of a selected set of preference criteria and more preferred for at least one of those.

\begin{mydef} (dominate)
\label{def-dominate}
Let $R=(A,D)$ and $R'=(A',D')$ be two repairs for CDP$(T, C, Or, M, W)$.
$R$ \emph{dominates} $R'$ with respect to a set of preference relations ${\cal
P}$ $\subseteq$ \{more complete, less incorrect, $\subset$\} if $R$ is more or equally preferred to $R'$ for all preference relations in ${\cal P}$ $\wedge$ $R$ is more preferred to $R'$ for at least one of the preference relations in ${\cal P}$.
\end{mydef}

Using the definition of dominate, we can now define a preference relation that combines the basic preference relations, but which prioritizes one of those. We prefer repairs that are preferred with respect to a prioritized basic preference relation, and that are not dominated by other such repairs. Def. \ref{def-X-optimal} formalizes this.

\begin{mydef} (combining with priority to one of the preference relations)
\label{def-X-optimal}
Let $X$ $\in$ \{more complete, less incorrect, $\subset$\}. Let ${\cal P}$ $\subseteq$ \{more complete, less incorrect, $\subset$\} $\setminus$ \{$X$\}.
A repair $R$ for CDP$(T, C, Or, M, W)$  is said to be \emph{X-optimal} with respect to ${\cal P}$ iff  $R$ is preferred with respect to $X$ and there is no other repair that is preferred with respect to $X$ and dominates $R$ with respect to ${\cal P}$.
\end{mydef}

We can also define a preference relation that combines basic preference relations, but where the basic preference relations have equal priority. In this case we prefer repairs that are not dominated by other repairs according to the selected basic preferences. Def. \ref{def-skyline} formalizes this.

\begin{mydef} (combining with equal priority)
\label{def-skyline}
A repair $R$ for CDP$(T, C, Or, M, W)$  is said to be \emph{skyline-optimal} with respect to ${\cal P}$ iff  there is no other repair that dominates $R$ with respect to ${\cal P}$.
\end{mydef}

We note that if a repair is X-optimal with respect to ${\cal P}$, then it is skyline-optimal with respect to ${\cal P}$$\bigcup$\{X\}.

As examples, for the CDP in Fig. \ref{CDP-example} we have that $R_4$ is $\subset$-optimal with respect to \{less incorrect\} and $R_3$ is $\subset$-optimal with respect to \{more complete\}.
Further, $R_1$  is less-incorrect-optimal with respect to \{more complete\} and more-complete-optimal with respect to \{less incorrect\}.

The advantage of maximally complete and more-complete-optimal repairs is that a maximal body of correct information is added to the ontology and for the latter without redundancy and/or with removing as much wrong information as possible. The advantage of minimally incorrect and less-incorrect-optimal repairs is that a maximal body of wrong information is removed from the ontology and for the latter without redundancy and/or with adding as much correct information as possible. Although these are the most attractive repairs, in practice it is not clear how to generate such repairs, apart from a usually infeasible brute-force procedure that checks the correctness of all axioms with the oracle. (Although a strategy can be devised to check all without asking the oracle for each axiom, the number of requests will still be large.) Repairs prioritizing subset minimality ensure that there is no redundancy. The advantage of removing redundant axioms is the reduction of computation time as well as the reduction of unnecessary user interaction. However, in some cases redundancy may be interesting. For instance, developers may want to have explicitly stated axioms in the ontologies even though they are redundant. This can happen, for instance, for efficiency reasons in applications or as domain experts have validated asserted axioms, these may be considered more trusted than derived axioms. Furthermore, focusing on redundancy may lead to less complete or more incorrect repairs.
Skyline-optimal is a relaxed criterion. When, for instance, ${\cal P}$ = \{more complete, less incorrect\}, then a skyline-optimal repair with respect to ${\cal P}$ is a preferred repair with respect to correctness for a certain level of completeness, or a preferred repair with respect to completeness for a certain level of correctness.  In practice, as it is not clear how to generate more-complete-optimal and less-incorrect-optimal repairs, a skyline-optimal repair may be the next best thing, and in some cases (e.g., Sect. \ref{sec-repair-incomplete-TBox}) it is easy to generate {\it a} skyline-optimal repair. However, in general, the difficulty lies in reaching as high levels of completeness and as low levels of incorrectness as possible.

\section{State of the art - ontologies}
\label{sec-state-of-the-art-ontologies}

Most of the current work has focused on the correctness or the completeness of ontologies, but very little work has dealt with both. However, a naive combination of a completion step and a debugging step does not necessarily lead to repairs for the combined problem. In this section we discuss current work.

\subsection{Correctness}
\label{sec-repair-inconsistent-incoherent-TBox}

When only dealing with repairing the inconsistency or incoherence of Tboxes (semantic defects), only wrong information is dealt with. Therefore, in Def. \ref{Def-repair-ontology}, $M = \emptyset$ and 
$A = \emptyset$. 
In most current approaches the domain expert is not included. This means that choices are made solely based on the logic and that correct axioms may be removed from the ontologies. Therefore, not all solutions may actually be repairs as defined in Def. \ref{Def-repair-ontology} as requirement (ii) may not be satisfied.

There is much work on repairing semantic defects. Most approaches are based on finding explanations or justifications for the defects using a glass-box or black-box approach \cite{KPSH05}. 
A glass-box approach is based on the internals of the reasoning algorithm of a description logic reasoner.  
A black-box approach uses a description logic reasoner as an oracle to determine answers to standard description logic reasoning tasks such as checking concept satisfiability or subsumption with respect to an ontology. 

A general approach for repairing incoherent ontologies is the following (adapted from  \cite{Sch05}). (For inconsistent ontologies we can use a similar approach.)
For a given set of unsatisfiable concepts for an ontology, compute the minimal explanations for the defects, i.e., the minimal reasons for the unsatisfiability of concepts. These minimal reasons for the unsatisfiability of a concept are sets of axioms and are called minimal unsatisfiability-preserving sub-TBoxes (MUPS)  or justifications for the unsatisfiability. We need to compute these MUPS or justifications for all unsatisfiable concepts. From these we can compute the minimal incoherence-preserving sub-TBoxes (MIPS) which are the smallest sets of axioms in the original Tbox that cause that TBox to be incoherent. To repair the incoherent TBox, we need to remove at least one axiom from each MIPS. We now define the notions in this general repairing approach formally.

The definition of MUPS is given in Def. \ref{Def-MUPS}. A MUPS in a consistent TBox can be seen as a justification (Def. \ref{Def-justification}) for an unsatisfiable concept. Indeed, if we instantiate $\psi$ in Def. \ref{Def-justification} with $P \sqsubseteq \bottom$ we obtain the MUPS for $P$. The definition of MIPS is given in Def. \ref{Def-MIPS}. 

\begin{mydef} (MUPS) \cite{Sch05}
Let $T$ be a TBox and $P$ be an unsatisfiable concept in $T$.
A set of axioms $T'$ $\subseteq$ $T$ is a minimal unsatisfiability-preserving sub-TBox (MUPS) if  $P$ is unsatisfiable in $T'$ and  $P$ is satisfiable in every sub-TBox $T''$ $\subsetneq$ $T'$.
\label{Def-MUPS}
\end{mydef}

\begin{mydef} (Justification) (similar to \cite{KPHS07})
Let $T$ be a consistent TBox and $T \models \psi$. A set of axioms $T'$ $\subseteq$ $T$ is a justification for  $\psi$ in  $T$ if $T' \models \psi$ and $\forall T'' \subsetneq T': T'' \not\models \psi$
\label{Def-justification}
\end{mydef}

\begin{mydef} (MIPS) \cite{Sch05}
Let $T$ be an incoherent TBox.
A TBox $T'$ $\subseteq$ $T$ is a minimal incoherence-preserving sub-TBox (MIPS) if  $T'$ is incoherent and every sub-TBox $T''$ $\subsetneq$ $T'$ is coherent.
\label{Def-MIPS}
\end{mydef}

As mentioned, to repair the incoherent TBox, we need to remove at least one axiom from each MIPS. Essentially, this means we should compute a hitting set (Def. \ref{Def-hitting-set}) of the set of MIPS and remove the hitting set from the TBox. In \cite{Sch05} these hitting sets are called pinpoints.
Complexity results regarding this problem are given in \cite{PS10-KR,PS10-ECAI,PS17}. 

\begin{mydef} (hitting set) (\cite{Rei87})
Let ${\cal T}$ be a collection of sets. A hitting set for  ${\cal T}$ is a set $H$ $\subseteq$ $\bigcup_{S \in {\cal T}} S$ such that $\forall S \in {\cal T}: H \cap S \neq \emptyset$.
\label{Def-hitting-set}
\end{mydef}

As an example, consider the TBox in Fig. \ref{CDP-example}. 
This TBox is incoherent with unsatisfiable concepts $P_1$ and $P_3$. The set of MUPSs for $P_1$ is \{\{$P_1 \sqsubseteq P_2$ (ax1), $P_1 \sqsubseteq \neg P_4$ (ax3), $P_2 \sqsubseteq P_4$ (ax4)\}, \{$P_1 \sqsubseteq P_3$ (ax2), $P_3 \sqsubseteq P_5$ (ax6), $P_3 \sqsubseteq P_6$ (ax7), $P_5  \sqsubseteq \forall s.P_8$ (ax9), $P_6 \sqsubseteq \exists s.\neg P_8$ (ax10)\}\} while the set of MUPSs for $P_3$ is \{\{$P_3 \sqsubseteq P_5$ (ax6), $P_3 \sqsubseteq P_6$ (ax7), $P_5  \sqsubseteq \forall s.P_8$ (ax9), $P_6 \sqsubseteq \exists s.\neg P_8$ (ax10)\}\}. The set of MIPSs is \{\{$P_1 \sqsubseteq P_2$ (ax1), $P_1 \sqsubseteq \neg P_4$ (ax3), $P_2 \sqsubseteq P_4$ (ax4)\}, \{$P_3 \sqsubseteq P_5$ (ax6), $P_3 \sqsubseteq P_6$ (ax7), $P_5  \sqsubseteq \forall s.P_8$ (ax9), $P_6 \sqsubseteq \exists s.\neg P_8$ (ax10)\}\}. A possible hitting set is \{$P_1 \sqsubseteq P_2$ (ax1), $P_3 \sqsubseteq P_5$ (ax6)\}. 

In general, there may be several hitting sets for the set of MIPS.
Different approaches use different heuristics for ranking the possible repairs.

The first tableau-based algorithm for debugging of an ontology was proposed in \cite{SC03,Sch05}. (For an overview of how a tableau-based reasoner works, see, e.g.,  \cite{BN03}.) The work was motivated by the development of the DICE (Diagnoses for Intensive Care Evaluation) terminology. A glass-box approach was used for an ${\cal ALC}$ reasoner.  The branches in the tableau-based reasoner were used to compute MUPS. The MIPS were computed by taking a subset-reduction of the union of all MUPSs, where the subset-reduction of a set S of TBoxes is the smallest sub-set of S such that for all TBoxes T in S there is a TBox T' in the subset-reduction that is a subset of T \cite{SC03}. 
Computing MUPS and MIPS for an unfoldable ${\cal ALC}$ TBox was shown to be in PSPACE, where an unfoldable Tbox is a TBox where the left-hand sides of the CGIs are atomic concepts and the right-hand-sides contain no reference (direct or indirect) to the defined atomic concept.

Computing hitting sets takes linear time for the non-minimal case while the problem is NP-complete for the minimal case \cite{Sch05}. This approach was implemented for unfoldable ${\cal ALC}$ TBoxes in the system MUPSter \cite{SHCvH07}. The tableau algorithm  in \cite{MLBP06} can be seen as an extension of this work. It computes maximally satisfiable sub-TBoxes and does not require individual steps for computing MUPS and applying the hitting set algorithm.
Also the approach in \cite{MLBP06} finds maximally coherent sub-Tboxes and presents an EXP-TIME algorithm for unfoldable ${\cal ALC}$ Tboxes.
The DION system \cite{SHCvH07} uses a bottom-up algorithm to compute MUPS. For an unsatisfiable concept P it computes two sets of axioms $\Sigma$ and $S$ such that P is satisfiable in $S$, but not in $\Sigma$ $\bigcup$ $S$. Then subsets $S'$ of $\Sigma$ are computed such that P is unsatisfiable in $S$ $\bigcup$ $S'$. By removing redundancy from these sets we obtain MUPS. For efficiency reasons not all sets of axioms are checked, but the search is guided by a relevance function, e.g., by using only axioms that are in some way relevant to the unsatisfiable concept.
In \cite{KPSH05} a glass-box technique is used to compute MUPS 
(called set of support in \cite{KPSH05}) 
for OWL ontologies (${\cal SHOIN}$). In \cite{KPHS07} a method was proposed to calculate all justifications of an unsatisfiable concept. Both a glass-box and black-box technique are presented for computing a single justification. The glass-box technique is an extension from  \cite{KPSH05}, while the black-box technique is based on an expansion stage where axioms are added to an initially empty set until a concept becomes unsatisfiable and a shrinking step where extraneous axioms are removed. Then, given an initial justification, a black-box method computes all justifications using a variation of the hitting set tree algorithm \cite{Rei87}. 
We note that computing all justifications in inconsistent ontologies is more difficult than computing all justifications in consistent ontologies (which are the ones referred to in Def. \ref{Def-justification}), a possible reason being that defects can be dealt with one at a time in consistent ontologies, but for inconsistent ontologies the only information that we have is that the ontology is inconsistent \cite{HPS09}.
The BEACON system \cite{AMIMPM16} implements an algorithm for $\elp$ ontologies based on a translation of the normalized Tbox (i.e., the TBox is first rewritten into a specific format) into Horn clauses and computing  minimal correction subsets of the clauses, which in their turn refer to repairs in the normalized Tbox.

As there may be different ways to repair the ontologies and as computing justifications can be expensive, different heuristics and optimization approaches have been proposed (e.g., \cite{JGHZ14}).
In \cite{Sch05} a heuristic is used stating that axioms appearing in more MIPSs are likely to be more erroneous. Therefore, axioms appearing in the most MIPSs are removed first (or, in other words, are first added to the hitting set).
In \cite{KPSC06} an arity-based heuristic is  used which is similar to the heuristic in  \cite{Sch05}. Further, \cite{KPSC06} introduces heuristics based on the impact on the ontology when an axiom is removed and based on test cases applied by a user, e.g., by specifying desired and undesired entailments, which may be seen as an oracle that has validated certain entailments a priori. They also propose to use provenance information about the axioms as well as information about how often the elements in the axioms are used in the other axioms in the ontology to rank the axioms. Reiter's hitting set algorithm is modified to take into account the axiom rankings. The approach is implemented in a prototype for a plug-in to SWOOP. In \cite{KPSH05} root concepts are repaired first. A root concept is an unsatisfiable concept for which a contradiction in its definition does not depend on the unsatisfiability of another concept. The other unsatisfiable concepts are then derived concepts. Repairing root concepts may automatically repair derived concepts. The idea of roots is used in \cite{MMV11} where the debugging is not restricted to unsatisfiable concepts, but to axioms that are unwanted. A variant of justification, called root justification, for a set of axioms U is defined as a set of axioms RJ that is a justification of at least one of the axioms in U and there is no justification of an axiom in U that is a proper subset of RJ. The authors show via experiments that the number of root justifications is usually lower than the number of justifications. The idea of roots is also used in the ORE system \cite{LB10} that implements a sound (all found dependencies are correct), but incomplete (not all dependencies are found) algorithm. 
In \cite{JQH09} a notion of relevance between axioms is defined and used to guide the computation of justifications.
Patterns explaining unsatisfiability are used in \cite{JGHZ11} to optimize finding MUPS. 

One of the few interactive approaches in debugging of ontologies is test-driven ontology debugging. In this approach queries are generated that are classified by an oracle as true (positive) or false (negative). Repairs that do not conform to the answers are discarded. Essentially, this is a strategy to guide a domain expert through the space of possible repairs to choose the repair that eventually is executed. 
An important issue in this approach is how to generate the queries to the oracle (e.g., \cite{SFFR12}). As shown in \cite{RS18}, there are many strategies, but none performs best for all cases. The right choice of strategy, however, is important as in some experiments the overhead for the oracle effort for the worst strategy with respect to the best strategy was over 250\%.
A system that implements test-driven ontology debugging is OntoDebug \cite{SRS18} which implements a number of earlier described methods for computing repairs, and guides the user 
using queries to find a final repair.

An approach based on the idea of truth maintenance systems is proposed in \cite{FMVN13} where a set of rules is defined that are used to compute consequences from the axioms in the ontology and explanations for unsatisfiable concepts and properties. The relatively light-weight description logic that is used should be sufficient for the representation of many learned (in contrast to manually developed) ontologies.

An approach that has not been proposed earlier, but that follows naturally from  the definitions and preferences of Sect. \ref{sec-repair}, is to use the oracle for the axioms in the MIPSs. For every MIPS, remove the axioms $\psi$ such that $Or$($\psi$) = true. If at least one of the MIPS becomes the empty set, then there is no repair unless we are willing to remove correct information. Assuming we have non-empty MIPSs after the removal of correct axioms, a hitting set would result in a repair. When redundancy is removed, we obtain a subset minimal repair. Another possibility, as we have checked the correctness using the oracle, is to use all remaining axioms in all MIPSs (as for these axioms $\psi$ we have that $Or$($\psi$) = false). This repair is less incorrect than the repairs obtained using hitting sets.

 There are also approaches that map the debugging problem into a revision problem (e.g., \cite{RW09,NRG12,FQZZ16}). A revision state \cite{NRG12} is a tuple of ontologies (${\cal O}$, ${\cal O}^{\models}$, ${\cal O}^{\not\models}$) where ${\cal O}^{\models}$ $\subseteq$ ${\cal O}$, ${\cal O}^{\not\models}$ $\subseteq$ ${\cal O}$, and ${\cal O}^{\models}$ $\cap$ ${\cal O}^{\not\models}$ = $\emptyset$. ${\cal O}^{\models}$ represents the wanted consequences of ${\cal O}$, while ${\cal O}^{\not\models}$ represents the unwanted consequences. In a complete revision state we also have that ${\cal O}$ = ${\cal O}^{\models}$ $\cup$ ${\cal O}^{\not\models}$. For a CDP,  ${\cal O}^{\not\models}$ could be initialized with $W$ (and when dealing with completion,  ${\cal O}^{\models}$ could be initialized with $M$).  The approach in \cite{NRG12} is an interactive method where questions are asked to an oracle to decide whether an axiom is correct or not, and then consequences are computed and revision states are updated iteratively. The decision on which questions to ask is based on the computation of an axiom impact measure. 
 In \cite{FQZZ16} a MIPS approach is used in the definition of the revision operator. In \cite{RW09} the authors show how  ontology debugging relates to theoretical aspects in revision and show, for instance, that axiom pinpointing is related to the problem of finding kernels in revision.


\subsection{Completeness}
\label{sec-repair-incomplete-TBox}

Most of the work on completing ontologies has dealt with completing the is-a structure of ontologies. An all-knowing oracle is often assumed. Therefore, in Def. \ref{Def-repair-ontology}, $\forall$ $\psi_m$ $\in$ $M$: $Or$($\psi_p$) = true, $W = \emptyset$ and  $D = \emptyset$.


There is not much work on the repairing of missing is-a structure. Most approaches just add the detected missing is-a relations. This conforms to the solution where $A$ = $M$. When $T \cup M$ is consistent and $\forall$ p $\in$ $M$: $Or$(p) = true, we are guaranteed that $M$ is a solution. In the case {\it all} missing is-a relations were detected in the detection phase, this is essentially all that can be done (except for removing redundancy, if so desired).
If not all missing is-a relations were detected -  and this is the common case - there are different ways to repair the ontology which are not all equally interesting and we can use the earlier defined preference relations. 

As these approaches do not deal with correctness, 
Def. \ref{def-more-correct} and \ref{def-max-correct}
are not used, and $\subset_D$ should be removed in Def. \ref{def-subset}. In Def. \ref{def-dominate} and \ref{def-skyline},  ${\cal P}$ = \{more complete, $\subset$\}. In Def. \ref{def-X-optimal}, 'less incorrect' should be removed. In this case, the semantically maximal solutions in \cite{WDL14} are a special case of the maximally complete repairs where only subsumption axioms between atomic concepts are used. Further, the X-optimal and skyline-optimal repairs combine only completeness and subset minimality.

Interactive solutions to this completion problem have been proposed for taxonomies \cite{LLT09,LL13,LWD15}, 
for $\el$ TBoxes  \cite{WDL14,LWD15} and for $\alc$ TBoxes \cite{LDI12}. All algorithms compute logically correct solutions which then need to be validated for correctness according to the domain by a domain expert. It is assumed that the axioms $M$ and $A$  represent subsumption between atomic concepts in the ontologies. 
The algorithms for taxonomies and  (normalized) $\el$ TBoxes (unified notation in \cite{LWD15}) require that 
$\forall$ m $\in$ $M$: Or(m) = true, and thus $M$ is a repair. The algorithms start with a first step that computes skyline-optimal repairs with respect to \{ more complete, $\subset$ \} for each missing is-a relation. This step is different for different representation languages of the TBox. For taxonomies the algorithm tries to find ways to repair a missing is-a relation $P_1 \sqsubseteq P_2$ by adding axioms of the form $P'_1 \sqsubseteq P'_2$ where $P_1 \sqsubseteq P'_1$ and $P'_2 \sqsubseteq P_2$. For $\el$, additionally, is-a relations of the form $\exists r.P_1 \sqsubseteq \exists r.P_2$ are repaired by repairing $P_1 \sqsubseteq P_2$. For $\elpp$ also role hierarchies and role inclusions need to be taken into account. 
Then the algorithms combine and modify these repairs into a single skyline-optimal repair for the whole set of missing is-a relations. Further, the algorithms repeat this process iteratively by solving new completion problems where the new $M$ is set to the added axioms in $A$ in the previous iteration. The union of the sets of added axioms of all iterations (with optionally removal of redundancy) is the final repair. It is shown that the skyline-optimal repairs (including the final repair if redundancy is removed) found during the iterations of the new completion problems are skyline-optimal repairs for the original completion problem that are equally or more complete than the repairs found in the first iteration.
Complexity results for the existence problem (does a repair exist?), relevance problem (does a repair containing a given axiom exist?) and necessity problem (do  all repairs contain a given axiom?) in general and with respect to different preferences are given for $\el$ and $\elpp$ in \cite{WDL14,LWD15}.
In \cite{LDI12} an approach is proposed for $\alc$ TBoxes by modifying a tableau-based reasoner. Repairs are found by closing leaf nodes in the completion graphs generated by trying to disprove missing is-a relations using the tableau reasoner. Open leaf nodes are closed by finding pairs of statements of the form $x:P$ and $x:\neg N$ and asserting then that $P \sqsubseteq N$. Additionally, the  same technique as for taxonomies is applied.

A non-interactive solution, i.e., without validation of an oracle, that is independent of the constructors of the description logic (e.g., tested with ontologies with expressivity up to ${\cal SHOIN(D)}$) is proposed in \cite{DWM17}. In contrast to the previous approaches where the repairs only contain subsumption axioms between existing concepts, this approach introduces justification patterns that can be instantiated with existing concepts or 'fresh' concepts. Further, the notion of justification pattern-based repairs is introduced  which are a kind of repairs that are subset-minimal. Methods for computing all justification patterns as well as justification-based repairs are given.

\subsection{Completeness and correctness}
\label{sec-repair-both}

There is very little work on dealing with both completeness and correctness. 
In \cite{LI13,IL13} two versions of the RepOSE system are presented that support debugging and completing the is-a structure of ontologies (and mappings between ontologies) in an iterative and interleaving way. 
Wrong information is removed by calculating justifications and allowing a user to mark wrong is-a relations.  Missing information is added using the interactive techniques in Sect. \ref{sec-repair-incomplete-TBox}. As the system always warns the user of influences of new additions or deletions on previous changes, the system can guarantee a repair if such exists, but it does not always guarantee a skyline-optimal solution.

\section{State of the art - ontology networks}
\label{sec-state-of-the-art-networks}

Completing and debugging of ontology networks has received more and more attention. Similar to single ontologies, also for networks the quality is dependent on the availability of domain experts, and completely automatic systems may reduce the quality \cite{PFSC13}.

Our definitions in Sect. \ref{sec-repair}  and \ref{sec-state-of-the-art-ontologies} can be used for ontology networks by creating a TBox from the network (i.e., it includes all axioms of all TBoxes from the ontologies and treats all mappings in all alignments in the network as axioms) and using this TBox in the definitions. It also follows that the techniques for single ontologies can be used for ontology networks.
However, in much of the current research the axioms in the ontologies in the network and the axioms in the alignments are distinguished and treated differently.

The field of ontology alignment \cite{ES07} deals with completeness of alignments (and thus only completion of the alignments, not of the ontologies in the networks). Many ontology alignment systems have been developed and overviews can be found in, e.g., \cite{KS03,Noy04,SE05,ES07,LST09,SE13,DILL17} and at the ontology matching web site (\url{http://www.ontologymatching.org}). 
Usually ontology alignment systems take as input two source ontologies and output an alignment. Systems can contain a pre-processing component that, e.g., partitions the ontologies into mappable parts thereby reducing the search space for finding candidate mappings. Further, a matching component uses matchers that calculate similarities between the entities from the different source ontologies or mappable parts of the ontologies. They often implement strategies based on linguistic matching, structure-based strategies, constraint-based approaches, instance-based strategies, strategies that use auxiliary information or a combination of these. Each matcher utilizes knowledge from one or multiple sources. Candidate mappings are then determined by combining and filtering the results generated by one or more matchers. Common combination strategies are the weighted-sum and the maximum-based strategies. The most common filtering strategy is the threshold filtering. Many systems output the found candidate mappings as an alignment. This approach is mainly a detection approach and the actual repairing is to add the alignment into the network. However, it is well-known that to improve the quality user validation is necessary and several systems allow for user interaction in the different steps of the alignment including validation (see, e.g., overview in \cite{DILL17}). Some systems also allow the addition of partial results to influence the computation of new results and thus a repair can lead to a new detection phase. 
Some systems introduce other components such as recommendation for the settings for the components in the system. A system that integrates all of these is discussed in \cite{LK17}.

Regarding correctness, most approaches deal with mapping repair where mappings rendering the network incoherent or inconsistent are removed.  Usually, the axioms in the ontologies are considered more trustworthy than the mappings and thus mappings are removed, rather than axioms in the ontologies. 
Although detection of defects can be different for different existing systems, justification-based techniques are often used for the repairing as in \cite{MST07}, and the Radon \cite{JHQHS09}, ALCOMO \cite{Meilicke11}, LogMap  \cite{JG11,JGZH12} and AgreementMakerLight \cite{SFPC15} systems. 
Other heuristics than the ones in Sect. \ref{sec-state-of-the-art-ontologies} could be used. For instance, the conservativity principle \cite{SJG14} states that the integrated ontology should not induce any change in the concept hierarchies of the input ontologies. In \cite{Meilicke11,JMCH13,SFPC15} the confidence values of the mappings are taken into account and in \cite{MST07} a semantic similarity measure between concepts in the mappings is used.

Similar to the case of ontologies, some approaches for ontology networks use a revision approach, e.g., \cite{MST09,QJH09,Euzenat15}. Usually,  the ontologies remain the same, but the set of mappings is revised.

An approach that distinguishes between axioms in the ontologies and in the alignments, but gives equal priority to them using approaches in Sect. \ref{sec-state-of-the-art-ontologies}  is discussed in \cite{IL13,LI13}.


\section{Opportunities}
\label{sec-opportunities}

In this paper we have defined a framework for completing and debugging ontologies and shown the state of the art in the field. It is clear that many research opportunities still exist.

\subsection{Theory and algorithms}

There are still challenges regarding the development of algorithms. Many approaches have been proposed regarding correctness, but finding (preferred) repairs in an acceptable time is still an issue. The current heuristics and optimization are almost all related to logical properties. However, this does not fit non-semantic defects. Furthermore, for semantic defects solutions may be proposed that remove correct statements while there could exist repairs that only remove wrong statements. In this case involving a domain expert in the generation and validation of repairs seems to be a way forward. There is relatively little work on dealing with completeness. There is still a need for new approaches and interactive systems. The current system that allows for user interaction deals with light-weight ontologies, while the work that allows for higher expressivity is non-interactive.
Even less work deals with completeness and correctness.
We need work on algorithms guaranteeing different kinds of preferred repairs. The current system dealing with both completion and debugging does help a user to find a repair, but it cannot even guarantee to find skyline-optimal repairs.

From the theoretical point of view, there is quite some work on complexity results for debugging and some for completion. However, there is a need for results regarding completion and the combination of debugging and completion as well as results for all cases regarding preferred repairs. In addition to results for finding repairs, there are also the questions related to the checking of the existence of repairs, the relevance of an axiom (is there a repair containing a given axiom?) and the necessity of an axiom (do all repairs add/delete a given axiom?).


The current formalization of repair uses an oracle that replies true or false for an axiom. This means that we assume that an oracle always answers, although the answer may be correct or wrong.  However, it is also possible that the oracle does not know an answer. In this case we may want to extend the formalization to allow the oracle to answer unknown. A prudent approach may not allow unknown axioms in the add set of a repair, but they could be allowed in the delete set. A credulous approach may allow unknown axioms in the add set, but not in the delete set. Further, preferences may be defined related to the use of unknown axioms.

In some cases, for instance, when there is a consensus about some concepts and their relations to each other, we may want to state that certain parts of the ontology are correct and should not be changed. This would then restrict repairing of defects to not include these parts. This can be handled by extending the formalization of the complete-debug problem by using the notion of background knowledge which represents information about the relevance or importance of parts of the ontology (\cite{HQ07}). In \cite{SFFR12} background knowledge is used to represent parts of the ontology that are asserted to be correct and therefore should not be changed in debugging. This is an example of the requirement called exploitation of context in \cite{HQ07}.

In this survey a repair consists of set of axioms that are added and a set of axioms that are deleted from the ontology (Def. \ref{Def-repair-ontology}). However, by removing axioms sometimes correct inferences are lost. Therefore, instead of removing complete axioms we may want to weaken an axiom or rewrite an axiom such that only the parts that caused a defect are removed. Ways to formalize what a part of an axiom is in this sense, together with debugging algorithms, are presented in \cite{HPS08,DQF14}. A method that rewrites the axioms in the ontology into simpler axioms and debug these is shown in \cite{KPSC06}. In \cite{LSPV08} parts of axioms responsible for unsatisfiability of concepts are traced. Further, lost inferences are calculated for atomic concepts. 
Possible axiom changes can be tagged as harmful when they do not solve unsatisfiability or introduce new unsatisfiability. The authors also introduce the notion of helpful changes where part of an axiom is removed, but other axioms are added to make up for lost inferences. 
An approach for dealing with defects in incoherent or inconsistent ontologies by changing axioms is axiom weakening where an axiom is replaced by another axiom that has fewer consequences subset-wise. Different ways to compute such weaker axioms are presented in \cite{BKNP18,TCGPPK18}.
While the current work deals with debugging, there may be cases for completion as well where we essentially would like to modify existing axioms.
In principle, all these cases are already covered by the current framework as a change in an axiom can be represented by the removal of the original axiom together with the addition of the changed axiom. However, this would require solutions for the full problem in Def. \ref{Def-repair-ontology} (which, as we have seen, are not available yet) and we may want to introduce new preference relations based on additions and deletions that reflect axiom changes.

In this survey we have focused on ontologies that do not contain instances. When instances are available these could be used in detection or repairing \cite{BGSS07,LB10,DDL17,Euzenat17}. For instance, using instances is one of the main ways to detect inconsistent ontologies.

\subsection{User support}
\label{sec-user-involvement}

From a practical point of view it is clear that to obtain the best results of the completion and debugging, domain experts need to be involved. However, the support of the current tools regarding user involvement is still lacking.
For instance, one of the outcomes in a study with users on ontology authoring is that debugging is difficult \cite{VBJS14}. Although ontology development systems may have explanation facilities,  debugging is still cumbersome and ontology developers use their own strategies, such as running a reasoner frequently when adding new axioms to the ontology to detect possible defects. The authors also state that, although good work has been done in ontology debugging, not that much has been integrated in ontology development tools. According to the study SWOOP had good debugging support,  Prot{\'e}g{\'e} had explanation facilities, WebProt{\'e}g{\'e} and the free edition of TopBraid did not have such support.

There are thus several opportunities for further work on user involvement in completion and debugging systems. One way forward is to consult the guidelines for ontology alignment systems that are also valid for the more general completion and debugging systems.
Recommendations for user support for ontology alignment systems regarding user interfaces (partly based on \cite{FS07}) as well as infrastructure and algorithms are given in \cite{ILA15}. The former include support for manipulation, inspection and explanation of mappings, while the latter include, among others, support for sessions, reduction of user interventions, collaboration, recommendations by the system, system configuration, debugging, trial execution and temporary decisions. 
In \cite{LDFIJLP19} the authors focus more specifically on user validation in ontology alignment and discuss issues related to  the profile of the user, the services of the alignment system, and its user interface. Recommendations for tool support for each aspect are given and the support in current systems is presented. For the user profile the authors discuss user expertise in terms of domain, technology and alignment systems. For services they discuss stage of involvement, feedback demand and feedback propagation. Finally, for the user interface issues regarding visualization and interaction are discussed. The paper reports that while there have been significant advances on the part of alignment systems in these areas, there are still key challenges to overcome such as reducing user workload, balancing informativeness with cognitive load and balancing user workload with user errors.
More generally, the field of visualization and interaction for ontologies and Linked Data has received more and more attention during the recent years and in \cite{ILLP19} the issues of cognitive support, user profiles and visual exploratory analysis are briefly discussed.

There are also some specific problems that have been noted in the debugging and completion area.

An important issue is that domain experts make mistakes and thus the oracle makes mistakes (see third case in Sect. \ref{sec-influence-oracle}). This has been reported often, and for instance, the study in \cite{RJSF19} states that questions to the oracle about statements that are true receive more reliable answers, that domain experts are sometimes overconfident, and that they consider themselves as imperfect.
Some research has discussed approaches to deal with this issue. For instance, in \cite{BS09} an approach (for ontologies with instances) is presented where the domain expert can request the history of the given answers, correct wrongly given answers and continue, while in \cite{RJSF19} a prediction model is developed for predicting oracle errors.
The impact of oracle errors on the effectiveness and efficiency of ontology alignment systems are shown in \cite{LDFIJLP19} as assessed in the Interactive Anatomy track of the Ontology Alignment Evaluation Initiative 2015 - 2018.

Another issue that has been mentioned is that some algorithms work on normalized versions of a TBox and therefore the results may not be that intuitive in terms of the original ontology. This requirement is called preservation of structure in \cite{HQ07}.

Further, in \cite{BS09} it was reported that domain experts  would want to be able to sometimes postpone their answers as an oracle, e.g., to have the time to check up some information or reflect more deeply.

More generally, completion and debugging should be integrated in every ontology development methodology, such that developers can detect and repair defects as soon as possible. One of the few that has different steps regarding completion and debugging within the general framework is an extension of the eXtreme Design Methodology \cite{DLB15}.

\section{Conclusion}
\label{sec-conclusion}

As semantically-enabled applications require high-quality ontologies, developing and maintaining as correct and complete as possible ontologies is an important, although difficult task in ontology engineering. A key step for guaranteeing a certain level of correctness and completeness is ontology debugging and completion. 

In this survey paper we have reviewed the state of the art in ontology debugging and completion where we have focused on the repairing step. We have done this by introducing a formalization for the completion and debugging problem which allowed us to review and discuss the state-of-the-art in this field in a uniform way. Using this formalization we show that, traditionally, debugging and completion have been tackled separately, we compared different approaches, and gained insights in the field and we point to new opportunities for further research to advance the field.

\section*{Acknowledgements}

We thank the anonymous referees for comments that have led to improvements of the paper
as well as Mina Abd Nikooie Pour, Ying Li and Chunyan Wang for proofreading and checking the examples.
This work has been financially supported by the Swedish e-Science Research Centre (SeRC), the Swedish Research Council (Vetenskapsr{\aa}det, dnr 2018-04147) and the Horizon 2020 project SPIRIT (grant agreement No 786993).

 \bibliographystyle{elsarticle-num} 
  \bibliography{main.bib}






\section*{Appendix}

In Fig. \ref{bio-example} (and visually in Fig. \ref{bio-example-graph}) we show an example of a complete-debug-problem inspired by 
the Galen ontology (\url{http://www.openclinical.org/prj_galen.html}) in the $\el$ language. 
For the ontology represented by the TBox we assume that we have detected two missing axioms Endocarditis  $\sqsubseteq$ PathologicalPhenomenon and  GranulomaProcess  $\sqsubseteq$ NonNormalProcess. Further, we have detected that we can infer the wrong axiom PathologicalProcess $\sqsubseteq$ GranulomaProcess  from the ontology.

\begin{figure}[tb]
\begin{center}
\begin{small}
\begin{tabular}{| l |} \hline
$C$ = \{ GranulomaProcess, CardioVascularDisease, PathologicalPhenomenon, \\
Fracture, Endocarditis, Carditis,  InflammationProcess, PathologicalProcess, \\
NonNormalProcess\} \\
\\
$T$ = \{ 
CardioVascularDisease $\sqsubseteq$ PathologicalPhenomenon, \\
Fracture $\sqsubseteq$ PathologicalPhenomenon, \\
$\exists$hasAssociatedProcess.PathologicalProcess $\sqsubseteq$ PathologicalPhenomenon, \\
Endocarditis $\sqsubseteq$ Carditis, \\
Endocarditis $\sqsubseteq$ $\exists$hasAssociatedProcess.InflammationProcess, \\
PathologicalProcess $\sqsubseteq$ NonNormalProcess, \\
PathologicalProcess $\sqsubseteq$
InflammationProcess,\\
InflammationProcess $\sqsubseteq$
GranulomaProcess 
\} \\
\\
$M$ = \{ 
Endocarditis  $\sqsubseteq$ PathologicalPhenomenon, \\
GranulomaProcess  $\sqsubseteq$ NonNormalProcess
\} \\
\\
$W$ = \{ 
PathologicalProcess $\sqsubseteq$ GranulomaProcess 
\} \\
\\
The following axioms are correct according to the domain, i.e., \\ $Or$ returns $true$ for: \\
GranulomaProcess  $\sqsubseteq$ InflammationProcess, \\
GranulomaProcess  $\sqsubseteq$ PathologicalProcess, \\
GranulomaProcess  $\sqsubseteq$ NonNormalProcess, \\
InflammationProcess $\sqsubseteq$ PathologicalProcess, \\
InflammationProcess $\sqsubseteq$ NonNormalProcess,\\
PathologicalProcess $\sqsubseteq$ NonNormalProcess,\\
CardioVascularDisease $\sqsubseteq$ PathologicalPhenomenon,\\
Fracture $\sqsubseteq$ PathologicalPhenomenon, \\
Endocarditis $\sqsubseteq$ PathologicalPhenomenon, \\
Endocarditis $\sqsubseteq$ Carditis, \\
Endocarditis $\sqsubseteq$ CardioVascularDisease, \\
Carditis $\sqsubseteq$ PathologicalPhenomenon, \\
Carditis $\sqsubseteq$ CardioVascularDisease, \\
$\exists$hasAssociatedProcess.PathologicalProcess $\sqsubseteq$ PathologicalPhenomenon, \\
$\exists$hasAssociatedProcess.InflammationProcess $\sqsubseteq$ PathologicalPhenomenon, \\
Endocarditis $\sqsubseteq$ $\exists$hasAssociatedProcess.InflammationProcess,\\
Endocarditis $\sqsubseteq$ $\exists$hasAssociatedProcess.PathologicalPhenomenon.\\
Note that for an oracle that does not make mistakes, \\
if  $Or$(P $\sqsubseteq$ Q) = true,  then also $Or$($\exists$r.P $\sqsubseteq$ $\exists$r.Q)=true.\\
and if $Or$(P $\sqsubseteq$ Q)=true, then also  $Or$(P $\sqcap$ O $\sqsubseteq$ Q)=true.\\
For other axioms P $\sqsubseteq$ Q with P, Q $\in$ $C$, $Or$(P $\sqsubseteq$ Q) = false.\\
\hline
\end{tabular}
\end{small}
\end{center}
\caption{$\el$ example. ($C$ is the set of atomic concepts in the ontology. $T$ is a TBox representing the ontology. $M$ is a set of missing axioms. $W$ is the set of wrong axioms. $Or$ is the oracle representing the domain expert.)}
\label{bio-example}
\end{figure}

\clearpage

\begin{figure}[tb] 
\centering
\includegraphics[width=1\textwidth]{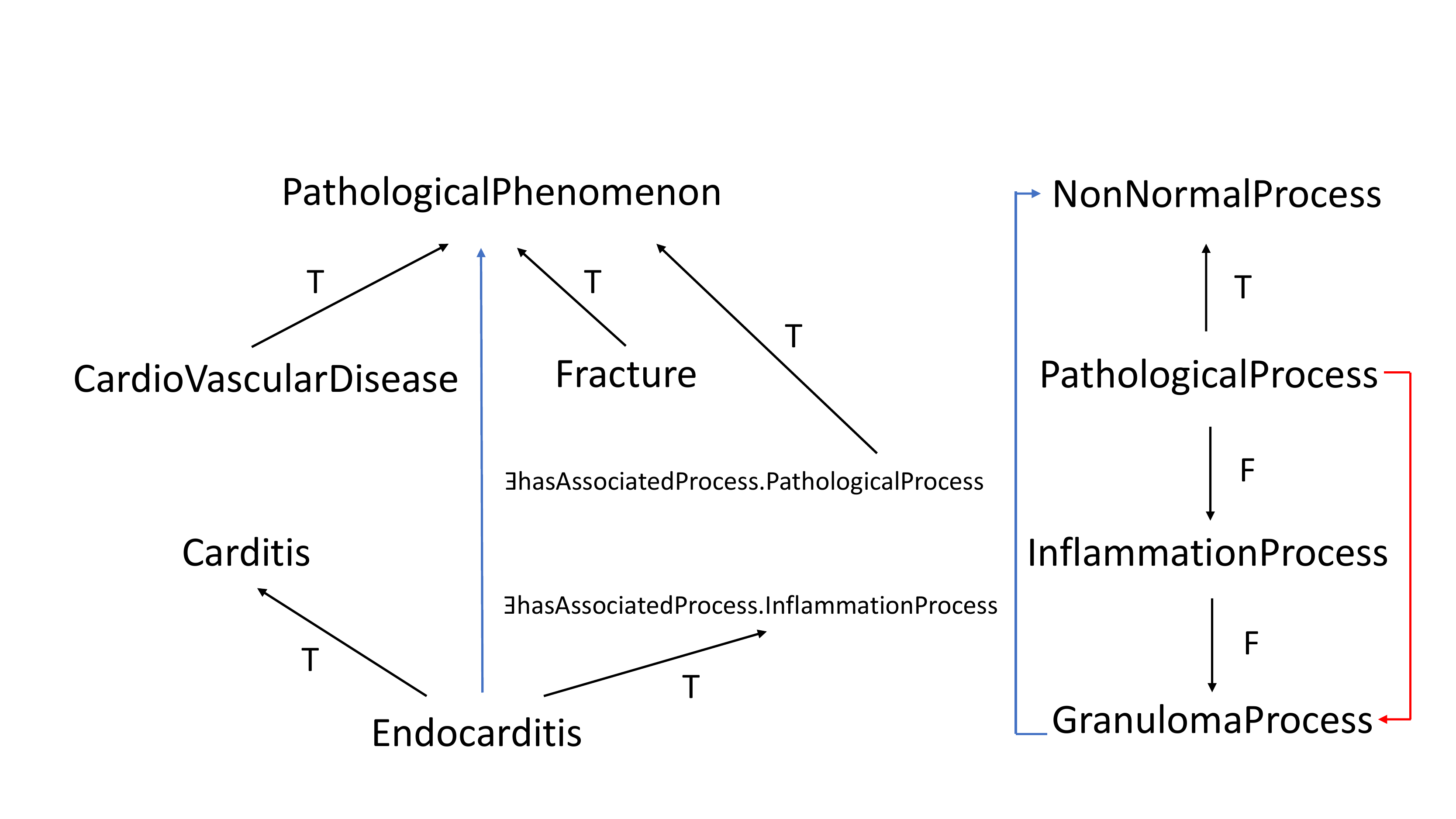}
\caption{Visualization of the example complete-debug problem in Fig. \ref{bio-example}. The axioms in the Tbox are represented with black arrows. The detected missing axioms are represented in blue. The detected wrong axiom is represented in red. The oracle's knowledge about the axioms in the ontology is marked with T (true) or F (false) at the arrows.} 
\label{bio-example-graph}
\end{figure}

\begin{figure}[tb] 
\centering
\includegraphics[width=1\textwidth]{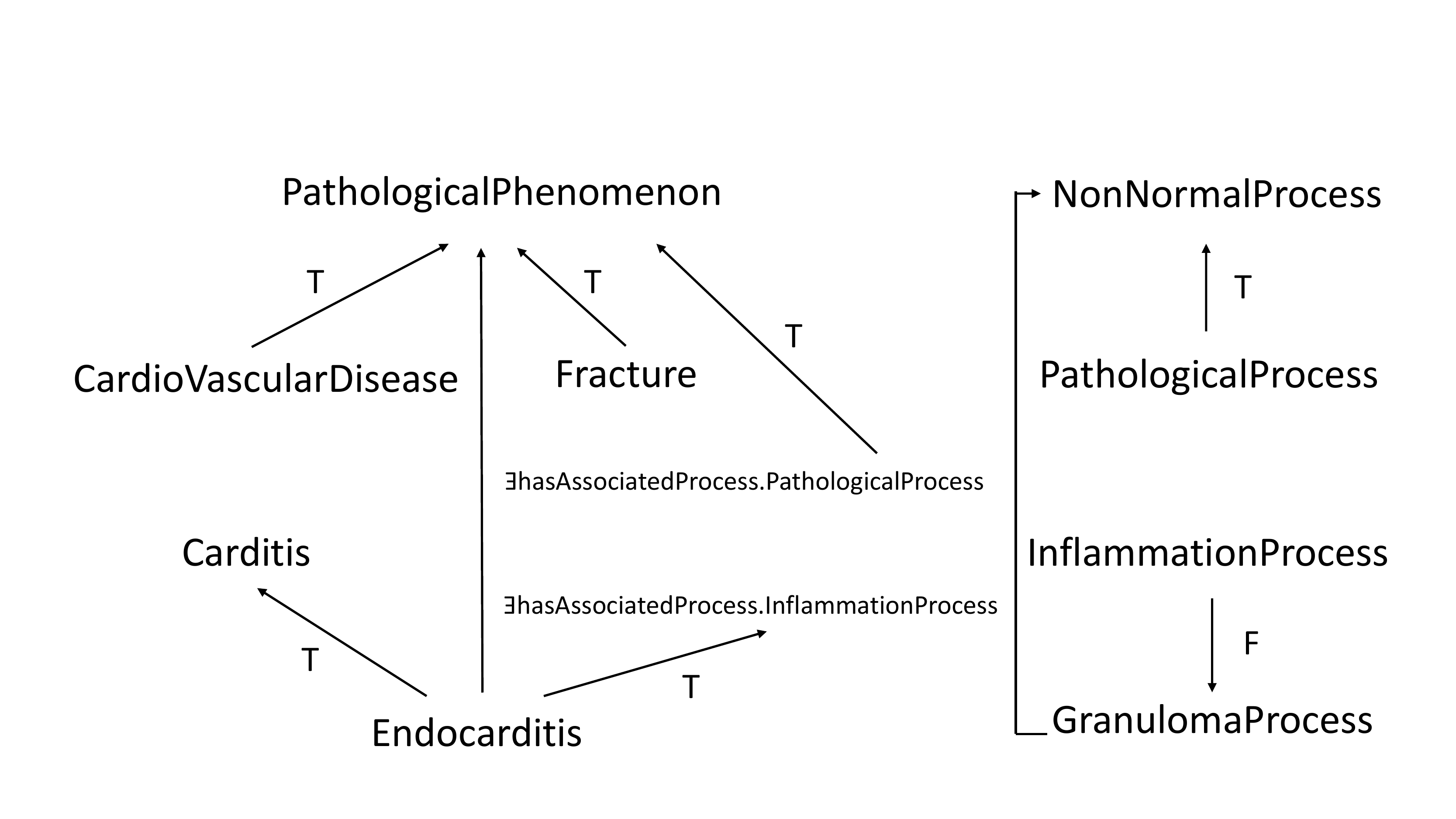}
\caption{Result after repairing the ontology in the complete-debug problem in Fig. \ref{bio-example} with repair R1. R1 = (A$_1$,D$_1$) with \\
A$_1$ = \{ Endocarditis  $\sqsubseteq$ PathologicalPhenomenon,\\
GranulomaProcess  $\sqsubseteq$ NonNormalProcess
\},
\\ D$_1$ =  \{ PathologicalProcess $\sqsubseteq$
InflammationProcess \}.} 
\label{bio-example-R1}
\end{figure}

\subsection*{Repairs}

There are different possible repairs of which we show some.
Fig. \ref{bio-example-R1}  shows the ontology after repair R1 is executed, where R1 = (A$_1$,D$_1$) with A$_1$ = \{ Endocarditis  $\sqsubseteq$ PathologicalPhenomenon,
GranulomaProcess  $\sqsubseteq$ NonNormalProcess
\} and D$_1$ =  \{ PathologicalProcess $\sqsubseteq$
InflammationProcess \}.
We show that R1 is a repair by discussing the different criteria in the definition of repair.
(i) Regarding the statements in A$_1$, we know that $Or$(Endocarditis  $\sqsubseteq$ PathologicalPhenomenon) = true, and 
$Or$(GranulomaProcess  $\sqsubseteq$ NonNormalProcess) = true. 
(ii) For the statement in D$_1$, we know that $Or$(PathologicalProcess $\sqsubseteq$ InflammationProcess) = false. 
(iii) $(T \cup A_1) \setminus D_1$ is consistent as $\el$ Tboxes are consistent.
(iv) $(T \cup A_1) \setminus D_1$ $\models$ Endocarditis  $\sqsubseteq$ PathologicalPhenomenon, as Endocarditis  $\sqsubseteq$ PathologicalPhenomenon has been explicitly added.
$(T \cup A_1) \setminus D_1$ $\models$ GranulomaProcess  $\sqsubseteq$ NonNormalProcess, as GranulomaProcess  $\sqsubseteq$ NonNormalProcess has been explicitly added.
(v) $(T \cup A_1) \setminus D_1$ $\not \models$ PathologicalProcess $\sqsubseteq$ GranulomaProcess.
The only justification of PathologicalProcess $\sqsubseteq$ GranulomaProcess  in the original ontology was 
\{ PathologicalProcess $\sqsubseteq$
InflammationProcess,
InflammationProcess $\sqsubseteq$
GranulomaProcess \} and this is not available anymore after the repair as PathologicalProcess $\sqsubseteq$ InflammationProcess is removed. The newly added axioms do not give rise to new justifications for PathologicalProcess $\sqsubseteq$ GranulomaProcess.

\begin{figure}[tb] 
\centering
\includegraphics[width=1\textwidth]{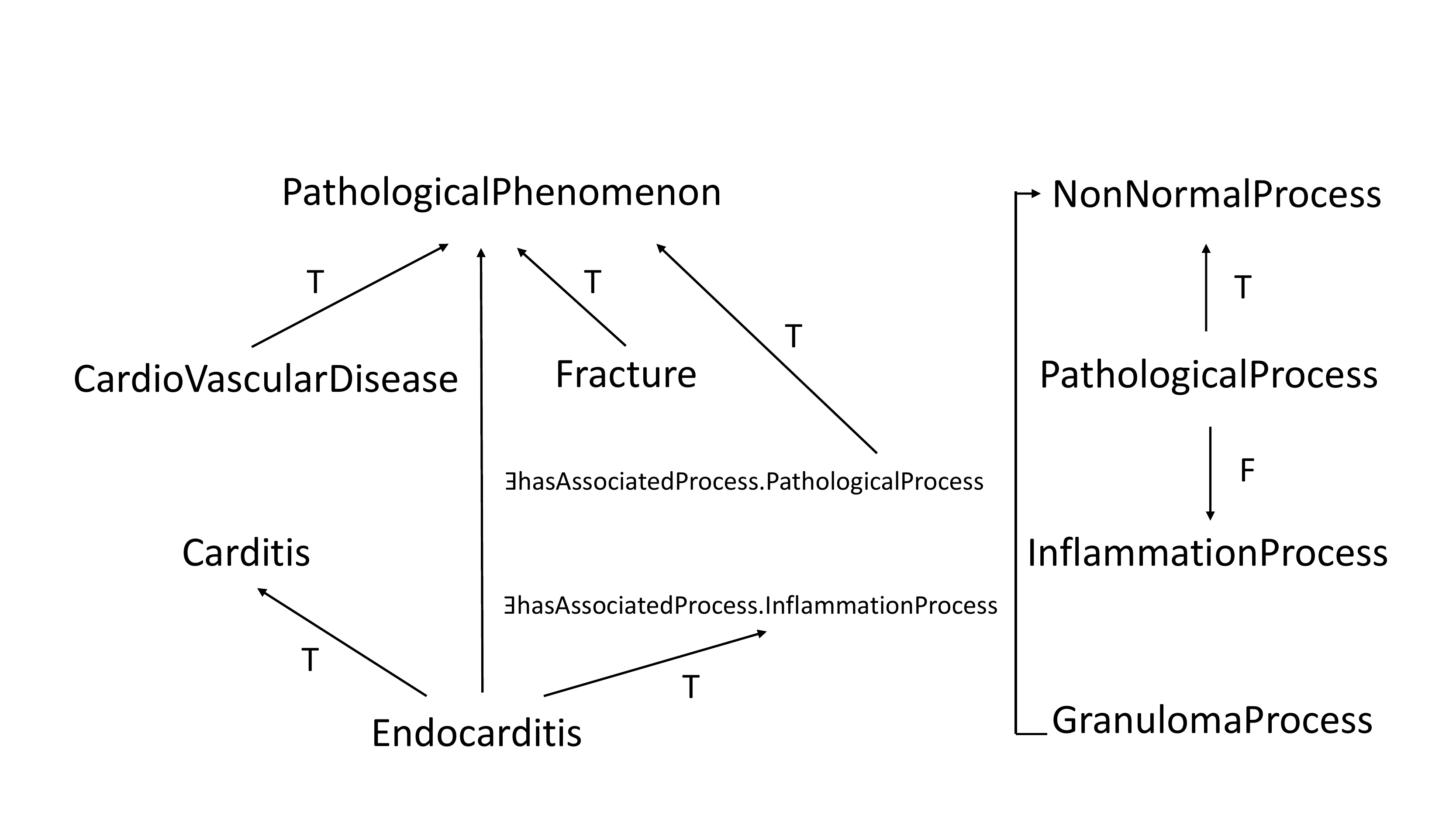}
\caption{Result after repairing the ontology in the complete-debug problem in Fig. \ref{bio-example} with repair R2. R2 = (A$_2$,D$_2$) with \\ 
A$_2$ =  \{ Endocarditis  $\sqsubseteq$ PathologicalPhenomenon,\\
GranulomaProcess  $\sqsubseteq$ NonNormalProcess
\},
\\ D$_2$ =  \{ InflammationProcess $\sqsubseteq$
GranulomaProcess \}.} 
\label{bio-example-R2}
\end{figure}

Fig. \ref{bio-example-R2}  shows the ontology after repair R2 is executed, where R2 = (A$_2$,D$_2$) with A$_2$ = A$_1$ = \{ Endocarditis  $\sqsubseteq$ PathologicalPhenomenon,
GranulomaProcess  $\sqsubseteq$ NonNormalProcess
\} and D$_2$ =  \{ InflammationProcess $\sqsubseteq$
GranulomaProcess  \}.
We show that R2 is a repair.
(i) Same as R1.
(ii) For the statement in D$_2$, we know that $Or$(InflammationProcess $\sqsubseteq$
GranulomaProcess) = false. 
(iii) Same as R1.
(iv) Same as R1.
(v) $(T \cup A_2) \setminus D_2$ $\not \models$ PathologicalProcess $\sqsubseteq$ GranulomaProcess.
The only justification of PathologicalProcess $\sqsubseteq$ GranulomaProcess  in the original ontology was 
\{ PathologicalProcess $\sqsubseteq$
InflammationProcess,
InflammationProcess $\sqsubseteq$
GranulomaProcess \} and this is not available anymore after the repair as InflammationProcess $\sqsubseteq$
GranulomaProcess is removed. The newly added axioms do not give rise to new justifications for PathologicalProcess $\sqsubseteq$ GranulomaProcess.

\clearpage

\begin{figure}[tb] 
\centering
\includegraphics[width=1\textwidth]{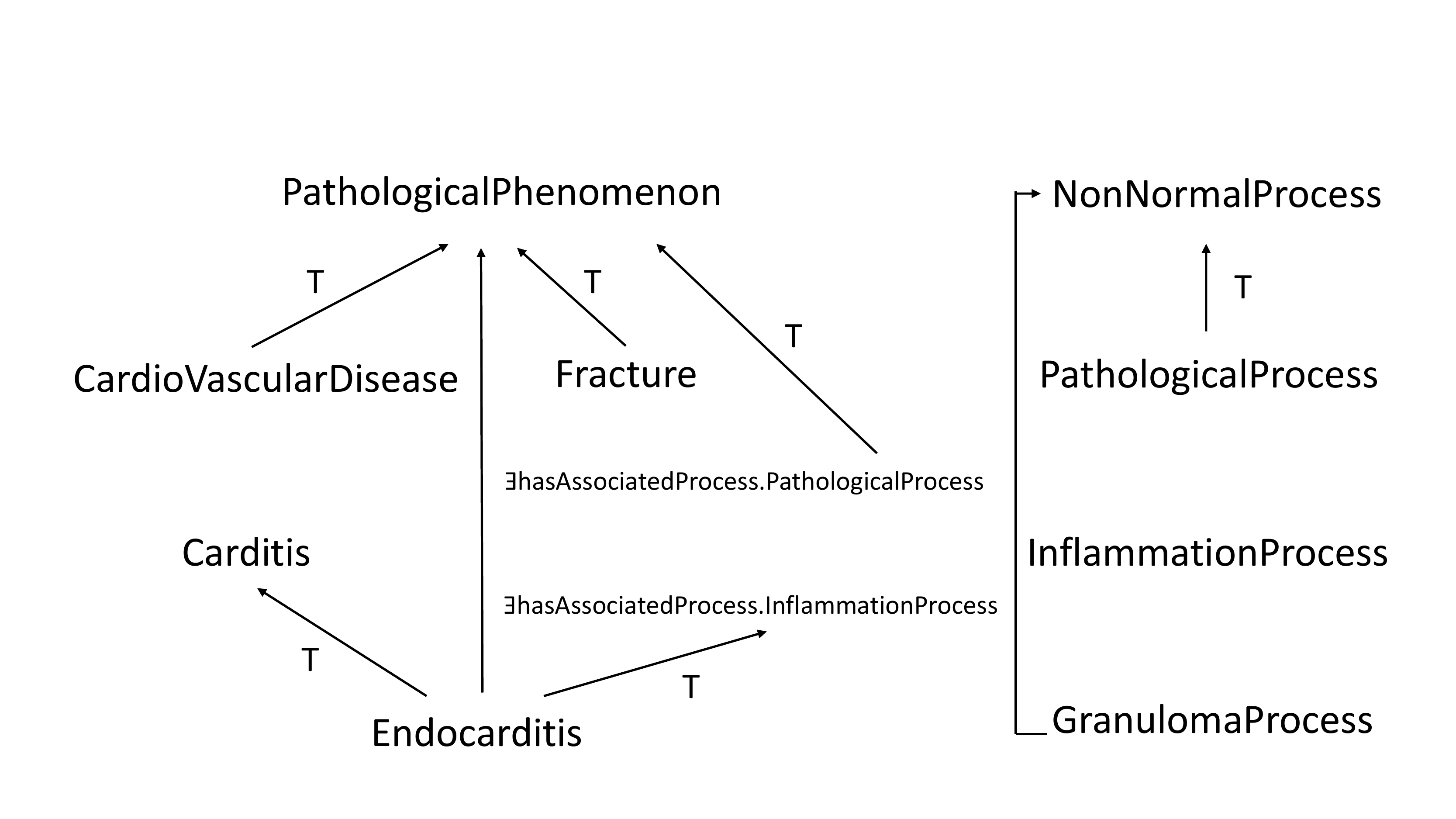}
\caption{Result after repairing the ontology in the complete-debug problem in Fig. \ref{bio-example} with repair R3. R3 = (A$_3$,D$_3$) with \\ A$_3$ = \{ Endocarditis  $\sqsubseteq$ PathologicalPhenomenon,\\
GranulomaProcess  $\sqsubseteq$ NonNormalProcess
\},  \\
D$_3$ =  \{ PathologicalProcess $\sqsubseteq$
InflammationProcess, \\
InflammationProcess $\sqsubseteq$
GranulomaProcess  \}.} 
\label{bio-example-R3}
\end{figure}

Fig. \ref{bio-example-R3}  shows the ontology after repair R3 is executed, where R3 = (A$_3$,D$_3$) with  A$_3$ = A$_1$ = A$_2$ = \{ Endocarditis  $\sqsubseteq$ PathologicalPhenomenon,
GranulomaProcess  $\sqsubseteq$ NonNormalProcess
\} and D$_3$ =  D$_1$ $\cup$ D$_2$ = \{ PathologicalProcess $\sqsubseteq$
InflammationProcess, InflammationProcess $\sqsubseteq$
GranulomaProcess  \}.
We show that R3 is a repair.
(i) Same as R1.
(ii) For the statements in D$_3$, we know that $Or$(PathologicalProcess $\sqsubseteq$ InflammationProcess) = false and $Or$(Inflammation\-Process $\sqsubseteq$
GranulomaProcess) = false. (See R1 and R2.)
(iii) Same as R1.
(iv) Same as R1.
(v) $(T \cup A_3) \setminus D_3$ $\not \models$ PathologicalProcess $\sqsubseteq$ GranulomaProcess.
The only justification of PathologicalProcess $\sqsubseteq$ GranulomaProcess  in the original ontology was 
\{ PathologicalProcess $\sqsubseteq$
InflammationProcess,
InflammationProcess $\sqsubseteq$
GranulomaProcess \} and this is not available anymore after the repair as the axioms in the justification have been removed. The newly added axioms do not give rise to new justifications for PathologicalProcess $\sqsubseteq$ GranulomaProcess.

\clearpage

\begin{figure}[tb] 
\centering
\includegraphics[width=1\textwidth]{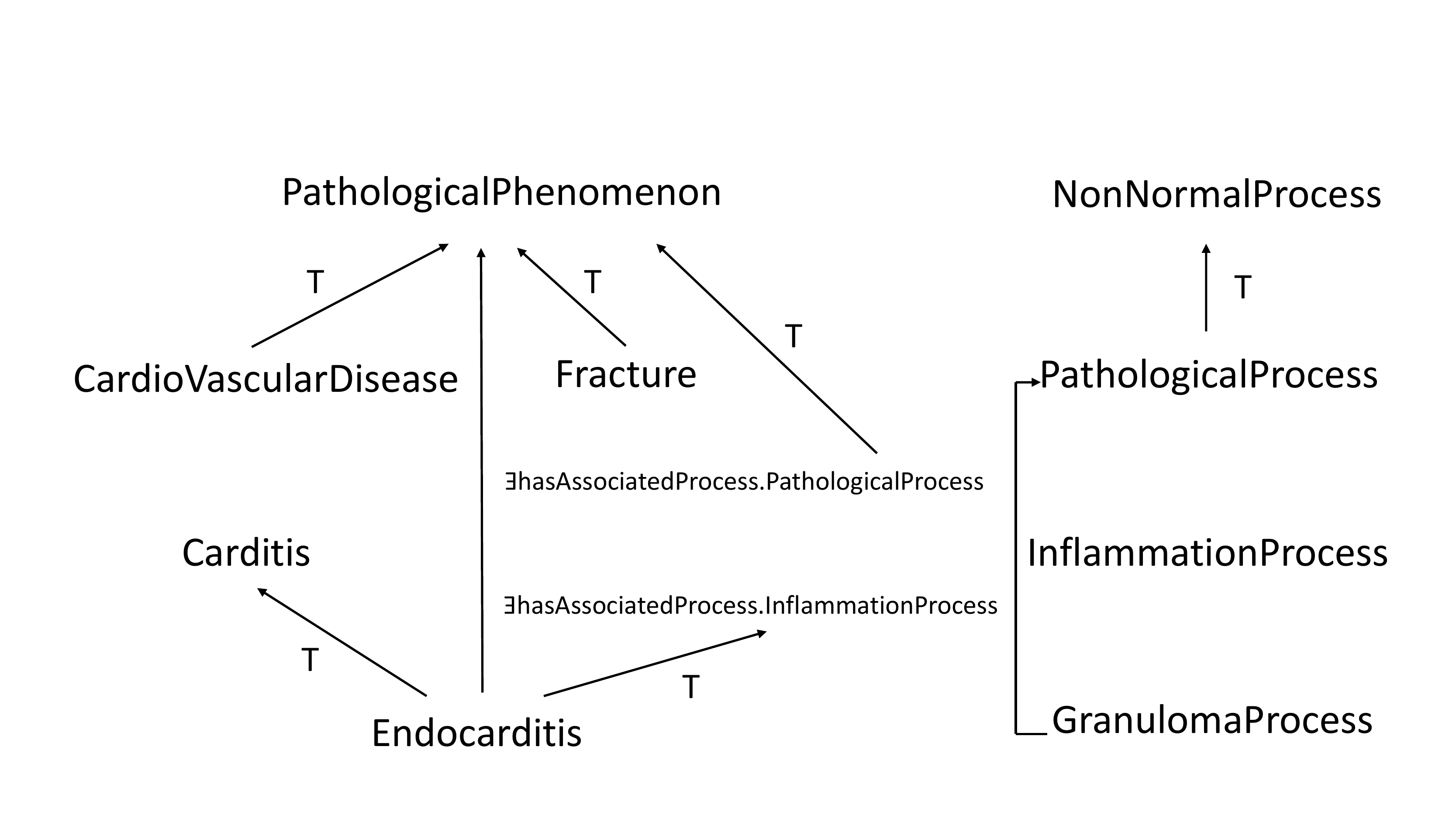}
\caption{Result after repairing the ontology in the complete-debug problem in Fig. \ref{bio-example} with repair R4. R4 = (A$_4$,D$_4$) with \\
A$_4$ = \{ Endocarditis  $\sqsubseteq$ PathologicalPhenomenon, \\
GranulomaProcess  $\sqsubseteq$ PathologicalProcess
\}, \\
D$_4$ =  \{ PathologicalProcess $\sqsubseteq$
InflammationProcess, \\
InflammationProcess $\sqsubseteq$
GranulomaProcess  \}.} 
\label{bio-example-R4}
\end{figure}

Fig. \ref{bio-example-R4}  shows the ontology after repair R4 is executed, where R4 = (A$_4$,D$_4$) with A$_4$ = \{ Endocarditis  $\sqsubseteq$ PathologicalPhenomenon,
GranulomaProcess  $\sqsubseteq$ PathologicalProcess
\} and D$_4$ = D$_3$ =  \{ PathologicalProcess $\sqsubseteq$
InflammationProcess, InflammationProcess $\sqsubseteq$
GranulomaProcess  \}.
We show that R4 is a repair.
(i) Regarding the statements in A$_4$, we know that $Or$(Endocarditis  $\sqsubseteq$ PathologicalPhenomenon) = true, and 
$Or$(GranulomaProcess  $\sqsubseteq$ PathologicalProcess) = true.  
(ii) Same as R3.
(iii) Same as R1.
(iv) $(T \cup A_4) \setminus D_4$ $\models$ Endocarditis  $\sqsubseteq$ PathologicalPhenomenon, as Endocarditis  $\sqsubseteq$ PathologicalPhenomenon has been explicitly added.
$(T \cup A_4) \setminus D_4$ $\models$ GranulomaProcess  $\sqsubseteq$ NonNormalProcess, as this can be derived from 
the added GranulomaProcess  $\sqsubseteq$ PathologicalProcess and the axiom PathologicalProcess  $\sqsubseteq$ NonNormalProcess that was already in the TBox.
(v) Same as R3.

\clearpage

\begin{figure}[tb] 
\centering
\includegraphics[width=1\textwidth]{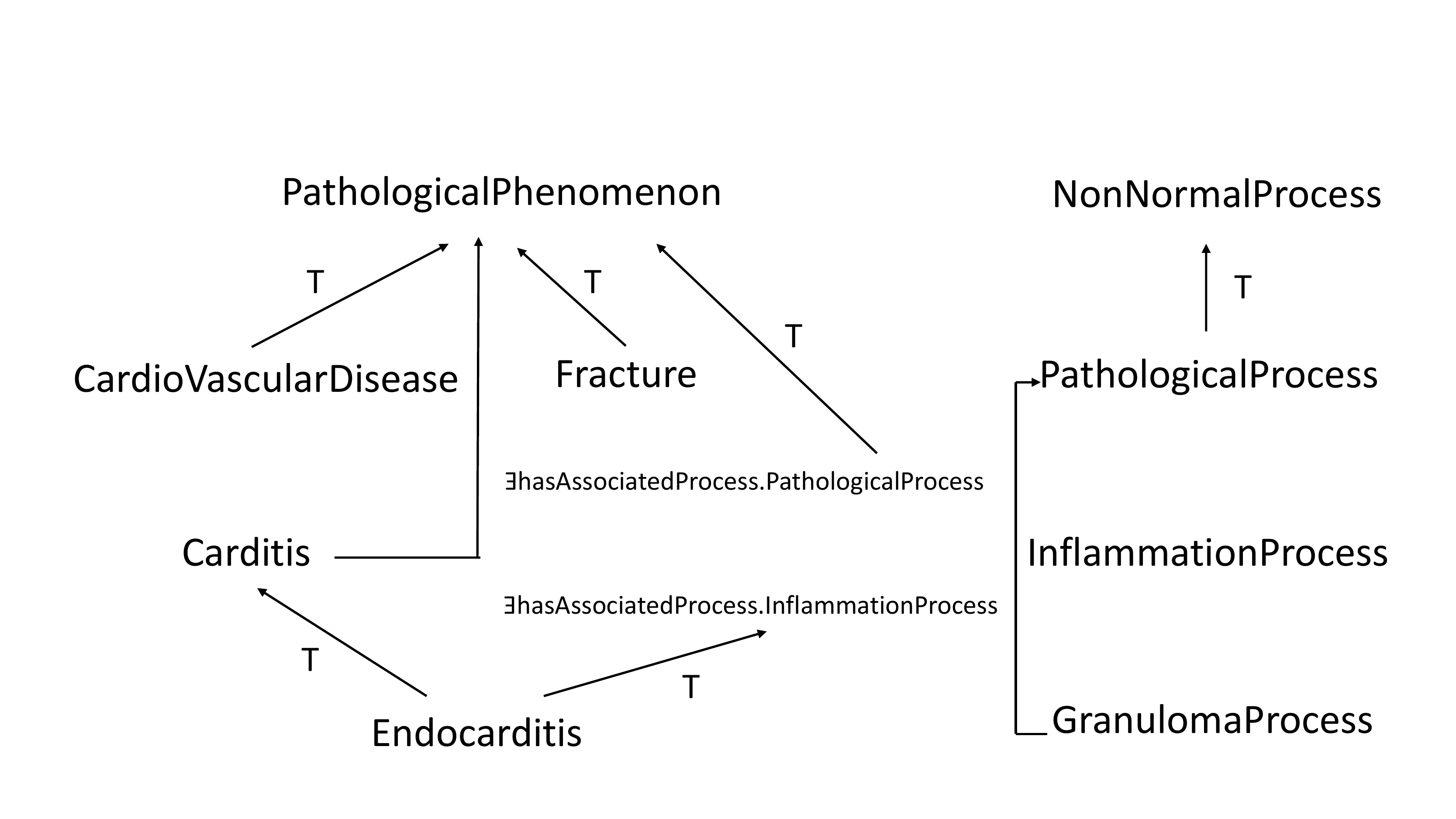}
\caption{Result after repairing the ontology in the complete-debug problem in Fig. \ref{bio-example} with repair R5. R5 = (A$_5$,D$_5$) with \\
A$_5$ = \{ Carditis  $\sqsubseteq$ PathologicalPhenomenon, \\
GranulomaProcess  $\sqsubseteq$ PathologicalProcess
\}, \\
D$_5$ =  \{ PathologicalProcess $\sqsubseteq$
InflammationProcess, \\
InflammationProcess $\sqsubseteq$
GranulomaProcess  \}.} 
\label{bio-example-R5}
\end{figure}

Fig. \ref{bio-example-R5}  shows the ontology after repair R5 is executed, where R5 = (A$_5$,D$_5$) with A$_5$ = \{ Carditis  $\sqsubseteq$ PathologicalPhenomenon,
GranulomaProcess  $\sqsubseteq$ PathologicalProcess
\} and D$_5$ = D$_3$ = D$_4$ = \{ PathologicalProcess $\sqsubseteq$
InflammationProcess, InflammationProcess $\sqsubseteq$
GranulomaProcess  \}.
We show that R5 is a repair.
(i) Regarding the statements in A$_5$, we know that $Or$(Carditis  $\sqsubseteq$ PathologicalPhenomenon) = true, and 
$Or$(GranulomaProcess  $\sqsubseteq$ PathologicalProcess) = true.  
(ii) Same as R3.
(iii) Same as R1.
(iv) $(T \cup A_5) \setminus D_5$ $\models$ Endocarditis  $\sqsubseteq$ PathologicalPhenomenon, as this can be derived from the axiom Endocarditis  $\sqsubseteq$ Carditis that was already in the TBox and
the added Carditis  $\sqsubseteq$ PathologicalPhenomenon. 
$(T \cup A_3) \setminus D_3$ $\models$ GranulomaProcess  $\sqsubseteq$ NonNormalProcess, for the same reason as in R4.
(v) Same as R3.

\clearpage

\begin{figure}[tb] 
\centering
\includegraphics[width=1\textwidth]{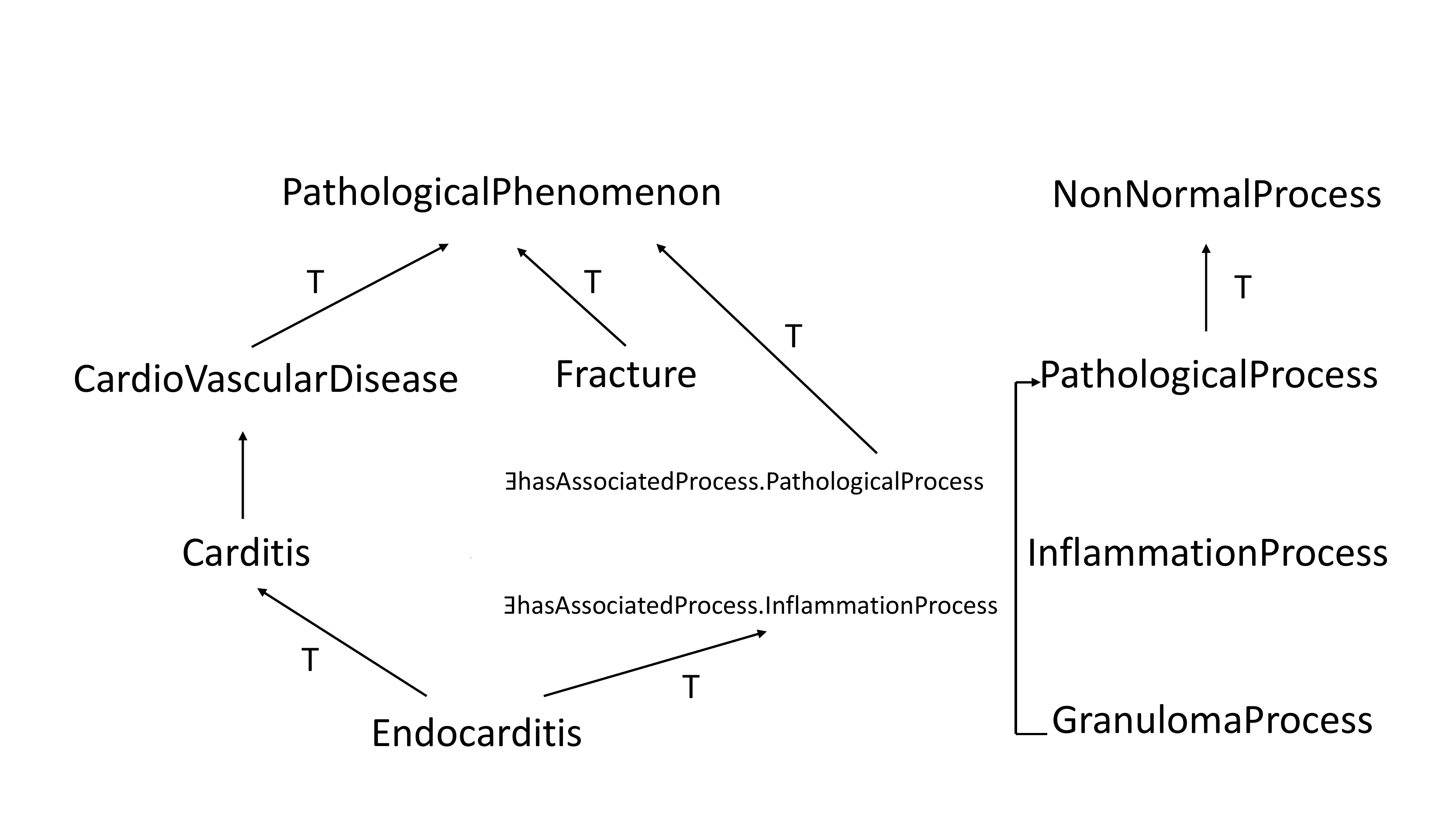}
\caption{Result after repairing the ontology in the complete-debug problem in Fig. \ref{bio-example} with repair R6. R6 = (A$_6$,D$_6$) with \\
A$_6$ = \{ Carditis  $\sqsubseteq$ CardioVascularDisease,\\
GranulomaProcess  $\sqsubseteq$ PathologicalProcess
\},
\\
D$_6$  =\{ PathologicalProcess $\sqsubseteq$
InflammationProcess, \\
InflammationProcess $\sqsubseteq$
GranulomaProcess  \}.} 
\label{bio-example-R6}
\end{figure}

Fig. \ref{bio-example-R6}  shows the ontology after repair R6 is executed, where R6 = (A$_6$,D$_6$) with A$_6$ = \{ Carditis  $\sqsubseteq$ CardioVascularDisease,
GranulomaProcess  $\sqsubseteq$ PathologicalProcess
\} and D$_6$ =  D$_3$ = D$_4$ = D$_5$ = \{ PathologicalProcess $\sqsubseteq$
InflammationProcess, InflammationProcess $\sqsubseteq$
GranulomaProcess  \}.
We show that R6 is a repair.
(i) Regarding the statements in A$_6$, we know that $Or$(Carditis  $\sqsubseteq$ CardioVascularDisease) = true, and 
$Or$(GranulomaProcess  $\sqsubseteq$ PathologicalProcess) = true.  
(ii) Same as R3.
(iii) Same as R1.
(iv) $(T \cup A_6) \setminus D_6$ $\models$ Endocarditis  $\sqsubseteq$ PathologicalPhenomenon, as this can be derived from the existing axiom Endocarditis  $\sqsubseteq$ Carditis,
the added axiom Carditis  $\sqsubseteq$ CardioVascularDisease and the existing axiom CardioVascularDisease $\sqsubseteq$ PathologicalPhenomenon.
$(T \cup A_6) \setminus D_6$ $\models$ GranulomaProcess  $\sqsubseteq$ NonNormalProcess, for the same reason as in R4.
(v) Same as R3.

\clearpage

\begin{figure}[tb] 
\centering
\includegraphics[width=1\textwidth]{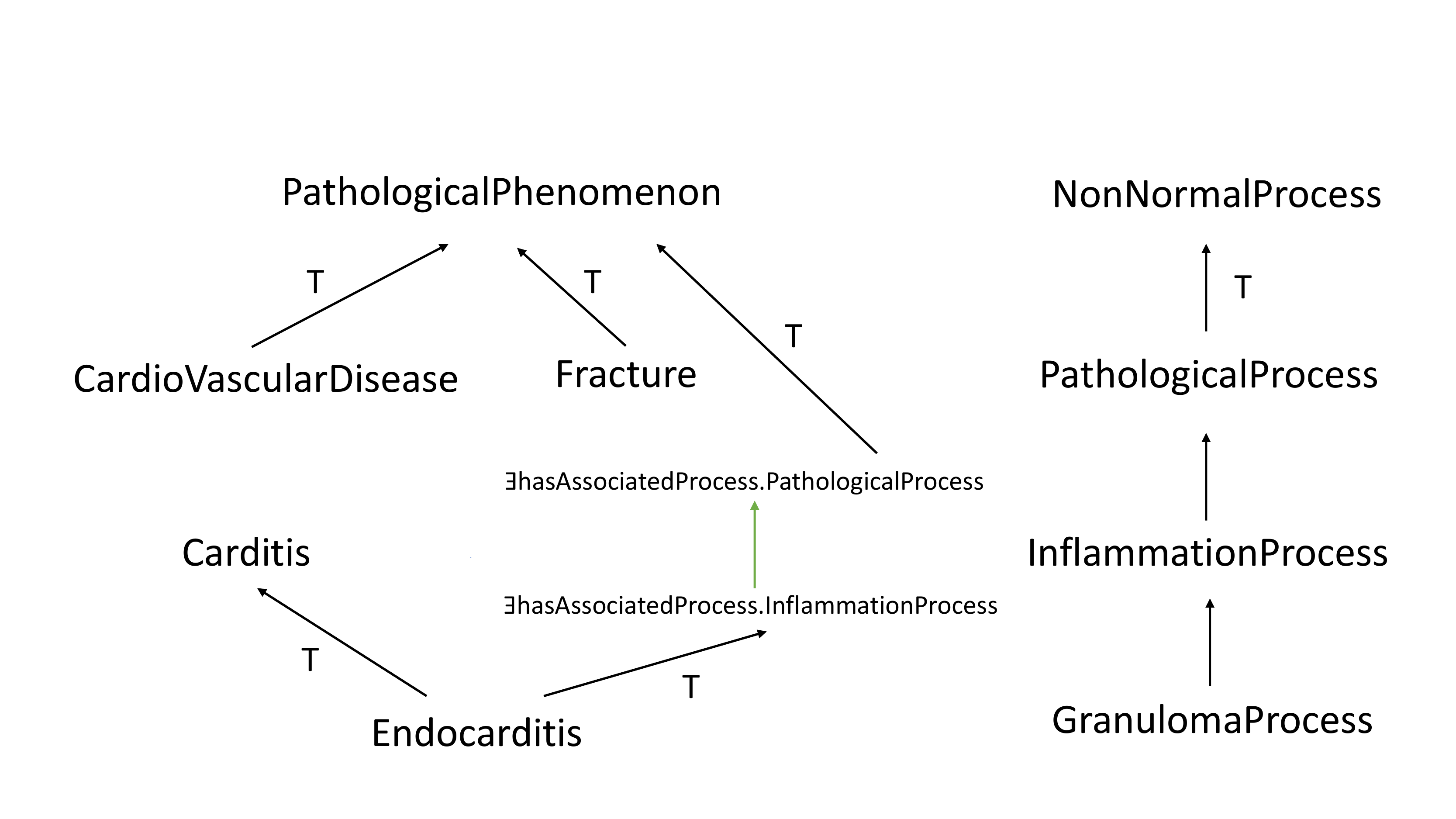}
\caption{Result after repairing the ontology in the complete-debug problem in Fig. \ref{bio-example} with repair R7. R7 = (A$_7$,D$_7$) with \\
A$_7$ = \{ GranulomaProcess  $\sqsubseteq$ InflammationProcess, \\
InflammationProcess $\sqsubseteq$ PathologicalProcess
\}, \\
D$_7$ =  \{ PathologicalProcess $\sqsubseteq$
InflammationProcess, \\
InflammationProcess $\sqsubseteq$
GranulomaProcess  \}.} 
\label{bio-example-R7}
\end{figure}

Fig. \ref{bio-example-R7}  shows the ontology after repair R7 is executed, where R7 = (A$_7$,D$_7$) with A$_7$ = \{ GranulomaProcess  $\sqsubseteq$ InflammationProcess, InflammationProcess $\sqsubseteq$ PathologicalProcess
\} and D$_7$ = D$_3$ = D$_4$ = D$_5$ = D$_6$ = \{ PathologicalProcess $\sqsubseteq$
InflammationProcess, InflammationProcess $\sqsubseteq$
GranulomaProcess  \}.
We show that R7 is a repair.
(i) Regarding the statements in A$_7$, we know that $Or$(GranulomaProcess  $\sqsubseteq$ InflammationProcess) = true, and 
$Or$(Inflammation\-Process $\sqsubseteq$ PathologicalProcess) = true.  
(ii) Same as R3.
(iii) Same as R1.
(iv) $(T \cup A_7) \setminus D_7$ $\models$ Endocarditis  $\sqsubseteq$ PathologicalPhenomenon, as this can be derived from the existing axiom Endocarditis $\sqsubseteq$ $\exists$hasAssociatedProcess.InflammationProcess, the newly derivable axiom $\exists$hasAssociatedProcess.InflammationProcess $\sqsubseteq$ $\exists$has\-AssociatedProcess.PathologicalProcess and the existing axiom
$\exists$hasAssociated\-Process.PathologicalProcess $\sqsubseteq$ PathologicalPhenomenon. The axiom  $\exists$has\-Associated\-Process.Inflammation\-Process $\sqsubseteq$ $\exists$hasAssociatedProcess.PathologicalProcess 
can be derived from InflammationProcess $\sqsubseteq$ PathologicalProcess which is newly added.
$(T \cup A_7) \setminus D_7$ $\models$ GranulomaProcess  $\sqsubseteq$ NonNormalProcess, as it can be derived from the two newly added axioms GranulomaProcess  $\sqsubseteq$ InflammationProcess and InflammationProcess $\sqsubseteq$ PathologicalProcess, together with the existing axiom PathologicalProcess $\sqsubseteq$ NonNormalProcess.
(v) Same as R3.

\begin{figure}[tb] 
\centering
\includegraphics[width=1\textwidth]{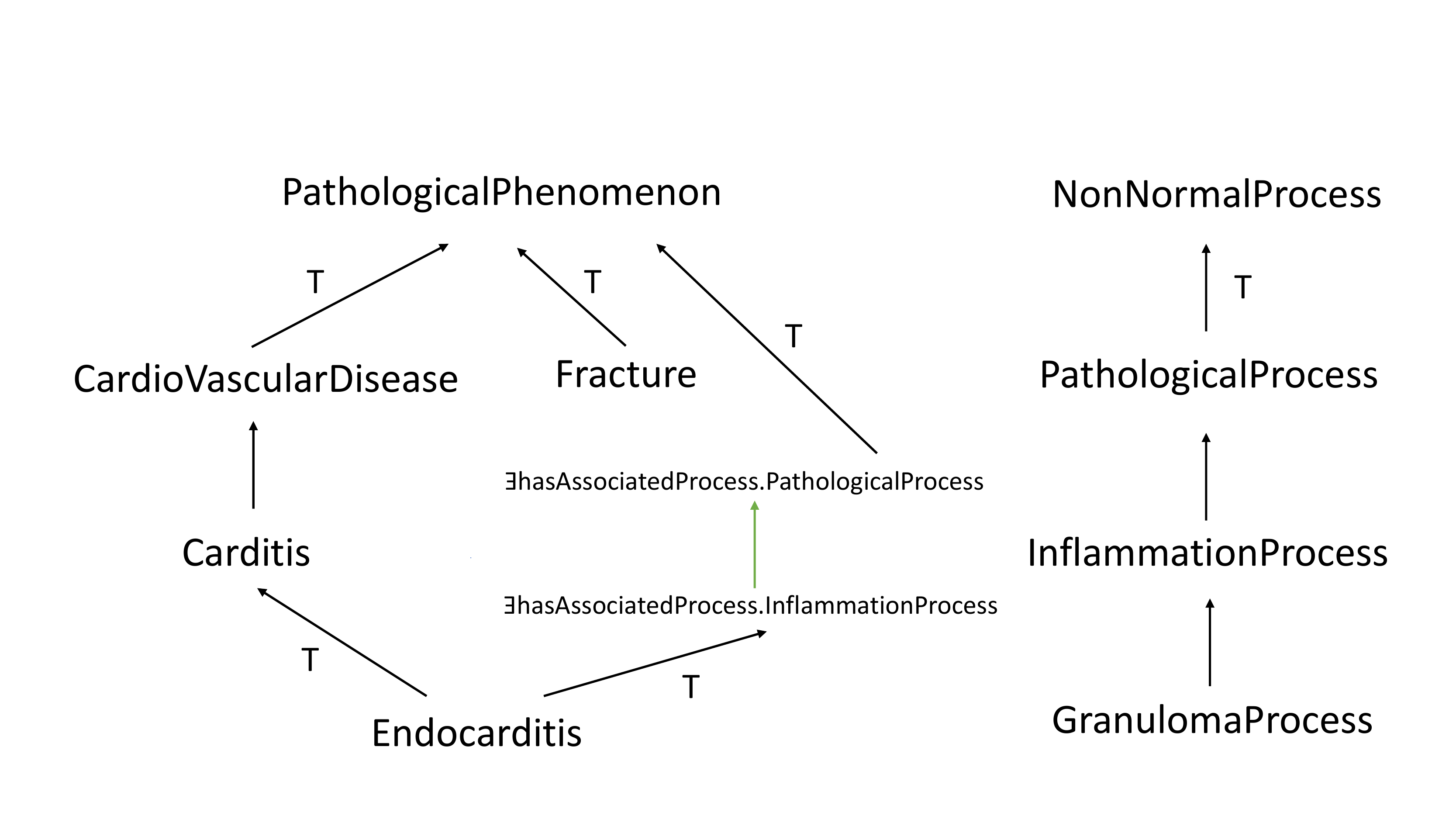}
\caption{Result after repairing the ontology in the complete-debug problem in Fig. \ref{bio-example} with repair R8. R8 = (A$_8$,D$_8$) with \\
A$_8$ = \{ Carditis  $\sqsubseteq$ CardioVascularDisease,\\ GranulomaProcess  $\sqsubseteq$ InflammationProcess,\\ InflammationProcess $\sqsubseteq$ PathologicalProcess
\}, \\
D$_8$ = \{ PathologicalProcess $\sqsubseteq$
InflammationProcess, \\
InflammationProcess $\sqsubseteq$
GranulomaProcess  \}.} 
\label{bio-example-R8}
\end{figure}

Fig. \ref{bio-example-R8}  shows the ontology after repair R8 is executed, where R8 = (A$_8$,D$_8$) with A$_8$ = \{ Carditis  $\sqsubseteq$ CardioVascularDisease, GranulomaProcess  $\sqsubseteq$ InflammationProcess, InflammationProcess $\sqsubseteq$ PathologicalProcess
\} and D$_8$ =  D$_3$ = D$_4$ = D$_5$ = D$_6$ =  D$_7$ = \{ PathologicalProcess $\sqsubseteq$
InflammationProcess, InflammationProcess $\sqsubseteq$
GranulomaProcess  \}.
We show that R8 is a repair.
(i) Regarding the statements in A$_8$, we know that
$Or$(Carditis  $\sqsubseteq$ CardioVascularDisease) = true,
$Or$(GranulomaProcess  $\sqsubseteq$ InflammationProcess) = true, and 
$Or$(InflammationProcess $\sqsubseteq$ PathologicalProcess) = true.  
(ii) Same as R3.
(iii) Same as R1.
(iv) $(T \cup A_8) \setminus D_8$ $\models$ Endocarditis  $\sqsubseteq$ PathologicalPhenomenon, as this can be derived from the existing axiom Endocarditis $\sqsubseteq$ $\exists$hasAssociatedProcess.InflammationProcess, the newly derivable axiom $\exists$has\-Associated\-Process.\-InflammationProcess $\sqsubseteq$ $\exists$hasAssociatedProcess.\-PathologicalProcess and the existing
$\exists$hasAssociatedProcess.PathologicalProcess $\sqsubseteq$ PathologicalPhenomenon. The axiom $\exists$hasAssociatedProcess.InflammationProcess $\sqsubseteq$ $\exists$hasAssociatedProcess.PathologicalProcess can be derived from InflammationProcess $\sqsubseteq$ PathologicalProcess which is newly added. (The missing axiom can also be derived from the existing axiom Endocarditis  $\sqsubseteq$ Carditis,
the added axiom Carditis  $\sqsubseteq$ CardioVascularDisease and the existing axiom CardioVascularDisease $\sqsubseteq$ PathologicalPhenomenon.)
$(T \cup A_8) \setminus D_8$ $\models$ GranulomaProcess  $\sqsubseteq$ NonNormalProcess, as it can be derived from the two newly added axioms GranulomaProcess  $\sqsubseteq$ InflammationProcess and InflammationProcess $\sqsubseteq$ PathologicalProcess, together with the existing axiom PathologicalProcess $\sqsubseteq$ NonNormalProcess.
(v) Same as R3.


\subsection*{Less incorrect}

R3, R4, R5, R6, R7 and R8 are less incorrect than R1 and R2, as R3, R4, R5, R6, R7 and R8 do not contain any wrong axioms but 
R1 contains the wrong axiom InflammationProcess $\sqsubseteq$
GranulomaProcess, and 
R2 contains the wrong axiom PathologicalProcess $\sqsubseteq$
InflammationProcess.
R3, R4, R5, R6, R7 and R8 are all equally incorrect. 

\subsection*{More complete}

Let R$_{ax}$ be the set of all correct axioms that can be derived from all repaired ontologies by repairs R1-R8 respectively. (If P $\sqsubseteq$ Q can be derived, then also  $\exists$r.P $\sqsubseteq$ $\exists$r.Q. If P $\sqsubseteq$ Q, then also  P $\sqcap$ O $\sqsubseteq$ Q. We do not add these statements nor tautologies.)
As none of these repairs has taken away a correct axiom from the ontology, all original correct axioms are in R$_{ax}$. Furthermore, all repairs need to make sure the axioms in $M$ are derivable and thus also these are in R$_{ax}$. Then, R$_{ax}$ = \{ CardioVascularDisease $\sqsubseteq$ PathologicalPhenomenon, 
Fracture $\sqsubseteq$ PathologicalPhenomenon, 
$\exists$hasAssociatedProcess.PathologicalProcess $\sqsubseteq$ PathologicalPhenomenon, 
Endocarditis $\sqsubseteq$ Carditis, 
Endocarditis $\sqsubseteq$ $\exists$hasAssociatedProcess.\-Inflam\-mation\-Process,
PathologicalProcess $\sqsubseteq$ NonNormalProcess,
GranulomaProcess $\sqsubseteq$ NonNormalProcess,
Endocarditis $\sqsubseteq$ PathologicalPhenomenon \}.

The set of axioms derivable from the ontology repaired by R1 is R1$_{ax}$ = R$_{ax}$ $\cup$ \{ InflammationProcess $\sqsubseteq$ NonNormalProcess \}. This additional axiom for the ontology repaired by R1 can be derived from (the wrong axiom) InflammationProcess  $\sqsubseteq$ GranulomaProcess and the added axiom GranulomaProcess $\sqsubseteq$ NonNormalProcess. 

For R2 and R3 the corresponding sets of axioms are R2$_{ax}$ = R3$_{ax}$ = R$_{ax}$, so no additional correct axioms.

R4$_{ax}$ = R$_{ax}$ $\cup$  \{ GranulomaProcess  $\sqsubseteq$ PathologicalProcess\}.

R5$_{ax}$ = R$_{ax}$ $\cup$ \{ GranulomaProcess  $\sqsubseteq$ PathologicalProcess, Carditis  $\sqsubseteq$ PathologicalPhenomenon \} = R4$_{ax}$  $\cup$ \{ Carditis  $\sqsubseteq$ PathologicalPhenomenon \}.

R6$_{ax}$ = R$_{ax}$ $\cup$  \{ GranulomaProcess  $\sqsubseteq$ PathologicalProcess,  Carditis  $\sqsubseteq$ PathologicalPhenomenon, Carditis  $\sqsubseteq$ CardioVascularDisease, Endocarditis  $\sqsubseteq$ CardioVascularDisease \} =  R5$_{ax}$ $\cup$ \{ Carditis  $\sqsubseteq$ CardioVascularDisease, Endocarditis  $\sqsubseteq$ CardioVascularDisease \}.

R7$_{ax}$ = R$_{ax}$ $\cup$ \{
GranulomaProcess  $\sqsubseteq$ InflammationProcess, 
GranulomaProcess  $\sqsubseteq$ PathologicalProcess,
InflammationProcess $\sqsubseteq$ PathologicalProcess, 
InflammationProcess $\sqsubseteq$ NonNormalProcess,
$\exists$hasAssociatedProcess.InflammationProcess $\sqsubseteq$ Pathological\-Phenomenon, 
Endocarditis $\sqsubseteq$ $\exists$hasAssociated\-Process.\-PathologicalPhenomenon
\} =
R1$_{ax}$  $\cup$ \{
GranulomaProcess  $\sqsubseteq$ InflammationProcess, 
GranulomaProcess  $\sqsubseteq$ PathologicalProcess,
InflammationProcess $\sqsubseteq$ PathologicalProcess,
$\exists$hasAssociated\-Process.\-InflammationProcess $\sqsubseteq$ PathologicalPhenomenon, 
Endocarditis $\sqsubseteq$ $\exists$has\-Associated\-Process.PathologicalPhenomenon \}

R8$_{ax}$ = R$_{ax}$ $\cup$ \{
GranulomaProcess  $\sqsubseteq$ InflammationProcess, 
GranulomaProcess  $\sqsubseteq$ PathologicalProcess,
InflammationProcess $\sqsubseteq$ PathologicalProcess, 
InflammationProcess $\sqsubseteq$ NonNormalProcess,
Carditis  $\sqsubseteq$ PathologicalPhenomenon, 
Carditis  $\sqsubseteq$ CardioVascularDisease,
Endocarditis  $\sqsubseteq$ CardioVascularDisease,
$\exists$hasAssociated\-Process.\-InflammationProcess $\sqsubseteq$ PathologicalPhenomenon, 
Endocarditis $\sqsubseteq$ $\exists$hasAssociated\-Process.\-PathologicalPhenomenon
\} = R6$_{ax}$ $\cup$ R7$_{ax}$

Thus,
R$_{ax}$ = R2$_{ax}$ = R3$_{ax}$ $\subsetneq$ R1$_{ax}$ $\subsetneq$ R7$_{ax}$ $\subsetneq$ R8$_{ax}$
and 
R$_{ax}$ = R2$_{ax}$ = R3$_{ax}$ $\subsetneq$ R4$_{ax}$ $\subsetneq$ R5$_{ax}$ $\subsetneq$ R6$_{ax}$  $\subsetneq$ R8$_{ax}$.
From this we conclude that R8 is more complete than R7 which is more complete than R1 which is more complete than R2 and R3. Further, R8 is more complete than R6 which is more complete than R5 which is more complete than R4 which is more complete then R2 and R3.

\subsection*{Subset}
For the add and delete sets for the repairs R1-R8 we obtain the following:\\
A$_1$ = A$_2$ = A$_3$, 
A$_7$ $\subsetneq$ A$_8$,
D$_1$ $\subsetneq$  D$_3$ = D$_4$ = D$_5$ = D$_6$ =  D$_7$ =  D$_8$, and 
D$_2$ $\subsetneq$  D$_3$ = D$_4$ = D$_5$ = D$_6$ =  D$_7$ =  D$_8$.
Therefore, R1 $\subset$ R3, R2 $\subset$ R3 and R7 $\subset$ R8. R3 deletes more wrong information than R1 and R2, respectively, but for repairing this is redundant (although R3 is less incorrect than R1 and R2). R8 adds additional correct axioms compared to R7, but this is redundant for the repairing (although R8 is more complete than R7).

\subsection*{Preferred repairs}

R8 is maximally complete as the ontology repaired by R8  contains all correct information.

R3, R4, R5, R6, R7 and R8 are minimally incorrect as the ontologies repaired by any of these repairs do not contain incorrect axioms.

R1 and R2 are subset minimal as if we remove an axiom from the add or delete set we would not have a repair anymore.
None of the other repairs is subset minimal as there are always variants where we can remove one of the axioms in the delete sets and still have repairs.

\subsection*{Combined preferences}

According to the definition, only preferred repairs with respect to a preference X can be X-optimal.

Regarding  more complete-optimal the only candidate among our example repairs is R8. R8 is more complete-optimal with respect to \{ less incorrect \} as the ontology repaired by R8 does not contain wrong information.  It is not more complete-optimal with respect to \{ $\subset$ \} as there are repairs that are also preferred with respect to more complete, but that remove one fewer wrong axiom. However, R8 is more complete-optimal with respect to \{ less incorrect, $\subset$ \}. If there would be a preferred repair with respect to more complete that dominates R8 with respect to \{ less incorrect, $\subset$ \}, then it would have to be more preferred to R8 with respect to $\subset$ as it cannot be more preferred with respect to less incorrect. However, removing an added axiom from R8 would make the repair not preferred with respect to more complete and removing fewer deleted axioms would make the repair less incorrect than R8.

Regarding less incorrect-optimal the candidates among our repairs are R3, R4, R5, R6, R7 and R8.
R8 is less incorrect-optimal with respect to \{ more complete \}. Similar reasoning as above leads to the fact that R8 is less incorrect-optimal with respect to \{ more complete, $\subset$ \}.
As R8 dominates R3, R4, R5, R6 and R7 with respect to more complete, R3, R4, R5, R6 and R7 cannot be less incorrect-optimal with respect to \{ more complete \}.
R7 dominates R8 with respect to \{ $\subset$ \}, so R8 cannot be less incorrect-optimal with respect to \{ $\subset$ \}.
A repair that would be preferred with respect to less incorrect and dominate R3, R4, R5, R6 or R7 with respect to \{ $\subset$ \} would need to remove the two wrong axioms (to be preferred with respect to less incorrect) and would therefore need to add fewer (subset-wise) axioms. However, removing one of the added axioms in R3, R4, R5, R6 and R7 would lead to sets of axioms that are not a repair. Therefore, R3, R4, R5, R6 and R7 are less incorrect-optimal with respect to \{ $\subset$ \}.

Regarding $\subset$-optimal the candidates are R1 and R2. R1 is more complete than R2, so R2 cannot be $\subset$-optimal with respect to \{ more complete \}. Also R1 is not $\subset$-optimal with respect to \{ more complete \} as there is another subset minimal solution that dominates R1 with respect to \{ more complete \} (e.g. same delete set as R1, but Carditis  $\sqsubseteq$ CardioVascularDisease in the add set instead of Endocarditis $\sqsubseteq$ PathologicalPhenomenon). For similar reasons R1 and R2 are not $\subset$-optimal with respect to \{ more complete, less incorrect \}.
They are, however, $\subset$-optimal with respect to \{ less incorrect \}. A less incorrect repair than R1 or R2 would need to remove both wrong axioms, but would then not be subset minimal.

We note that all preferred repairs are skyline optimal. Further, if a repair is X-optimal with respect to ${\cal P}$, then it is skyline-optimal with respect to ${\cal P}$ $\bigcup$ $X$. 
Thus, R8 is skyline-optimal with respect to \{ more complete, less incorrect \} and \{ more complete, less incorrect, $\subset$ \}.
R1, R2, R3, R4, R5, R6 and R7 are skyline-optimal with respect to \{ less incorrect, $\subset$ \}.

In addition there are also skyline-optimal repairs that are not X-optimal. For instance, 
R1 is not more-complete-optimal with respect to \{  $\subset$ \} nor $\subset$-optimal  with respect to \{ more complete \}.
However, R1 is skyline-optimal with respect to \{ more complete, $\subset$ \}. If there would be a repair that dominates R1 with respect to \{ more complete, $\subset$ \}, then there are two possibilities. The first possibility is that the other repair is more preferred with respect to $\subset$, which would mean taking away an axiom in the add set or in the delete set, but then we do not have a repair. The second possibility is that the other repair is more preferred with respect to more complete and equally preferred with respect to $\subset$. The second condition would only be satisfied if the add and delete sets are the same, but then we have the same repair.

\end{document}